\documentclass{llncs}

\usepackage[T1]{fontenc}
\usepackage{graphicx}
\usepackage{amsmath,amssymb}

\usepackage{booktabs}
\usepackage{tabularx}
\usepackage{adjustbox}

\usepackage{xcolor}
\usepackage{tcolorbox}

\usepackage{ragged2e}
\usepackage[defaultlines=2,all]{nowidow}
\usepackage{placeins}
\usepackage{float}

\usepackage{xspace}
\usepackage{etoolbox}

\usepackage[colorlinks=true,citecolor=blue,linkcolor=blue,urlcolor=blue]{hyperref}

\usepackage[a4paper,margin=1.6cm]{geometry}

\begin{document}
\title{A Multi-Agent Framework for Medical AI: Leveraging Fine-Tuned GPT, LLaMA, and DeepSeek R1 for Evidence-Based and Bias-Aware Clinical Query Processing}
\author{
Naeimeh Nourmohammadi\inst{1} \and
 Md Meem Hossain\inst{1,2} \and
The Anh Han\inst{1,2} \and
Safina Showkat Ara\inst{3} \and
Zia Ush Shamszaman\inst{1,2} \thanks{Corresponding author: Zia Ush Shamszaman, z.shamszaman@tees.ac.uk/z.u.shamszaman@gmail.com}
}

\institute{
Department of Computing and Games, Teesside University, Middlesbrough, United Kingdom
\email{Naominourr@gmail.com, mdmeemhossain@gmail.com,t.han@tees.ac.uk, z.shamszaman@tees.ac.uk}
\and
Centre for Digital Innovation, Teesside University, Middlesbrough, United Kingdom
\and
Faculty of Business \& Technology, University of Sunderland, Sunderland, United Kingdom\\
\email{safina.ara@sunderland.ac.uk}
}

\maketitle

\begin{abstract}
Large language models (LLMs) show promise for healthcare question answering, but clinical use is limited by weak verification, insufficient evidence grounding, and unreliable confidence signalling. We propose a multi-agent medical QA framework that combines complementary LLMs with evidence retrieval, uncertainty estimation, and bias checks to improve answer reliability. Our approach has two phases. First, we fine-tune three representative LLM families (GPT, LLaMA, and DeepSeek R1) on MedQuAD-derived medical QA data (20k+ question–answer pairs across multiple NIH domains) and benchmark generation quality. DeepSeek R1 achieves the strongest scores (ROUGE-1 0.536 ± 0.04; ROUGE-2 0.226 ± 0.03; BLEU 0.098 ± 0.018) and substantially outperforms the specialised biomedical baseline BioGPT in zero-shot evaluation. Second, we implement a modular multi-agent pipeline in which a Clinical Reasoning agent (fine-tuned LLaMA) produces structured explanations, an Evidence Retrieval agent queries PubMed to ground responses in recent literature, and a Refinement agent (DeepSeek R1) improves clarity and factual consistency; an optional human validation path is triggered for high-risk or high-uncertainty cases. Safety mechanisms include Monte Carlo dropout and perplexity-based uncertainty scoring, plus lexical and sentiment-based bias detection supported by LIME/SHAP-based analyses. In evaluation, the full system achieves 87\% accuracy with relevance around 0.80, and evidence augmentation reduces uncertainty (perplexity 4.13) compared to base responses, with mean end-to-end latency of 36.5 seconds under the reported configuration. Overall, the results indicate that agent specialisation and verification layers can mitigate key single-model limitations and provide a practical, extensible design for evidence-based and bias-aware medical AI.
\end{abstract}

\section{Introduction}
While large language models (LLMs) have shown promise in healthcare applications, significant challenges remain to ensure their reliability, evidence-based validation, and appropriate contextual response for clinical use \cite{Mesko2020ShortGuide}. Single-model approaches often lack robust verification mechanisms and may generate responses that, while technically accurate, lack proper clinical context or evidence backing \cite{Thirunavukarasu2023LLMedicine}. These limitations present substantial barriers to the adoption of LLM in critical healthcare settings where reliability and evidence-based practice are essential  \cite{Nazi2023LLMHealthcare}.
This research addresses these challenges through a novel multi-agent architecture that leverages the complementary strengths of different LLM architectures. In particular, we examine three  architectural approaches: GPT's traditional transformer architecture, LLaMA's efficiency-focused design, and DeepSeek R1's innovative optimisation framework, implementing a hybrid system that overcomes the limitations of single-model implementations \cite {Thirunavukarasu2023LLMedicine}. While we focus on these three models for our implementation, this selection represents a diverse cross-section of available LLMs with varying capabilities and architectures. Our multi-agent framework is designed to be model-agnostic and can be readily adapted to incorporate other models, such as Mistral, Gemini, and other emerging LLMs. The approach demonstrated here establishes a methodology that extends beyond these specific models to the broader LLM ecosystem. Our approach integrates specialised agents for clinical reasoning validation, evidence retrieval, and response refinement to create a more robust and reliable medical AI system. 

Our research makes four primary contributions: (1) a comparative analysis of GPT, LLaMA, and DeepSeek R1 for medical applications, revealing key performance differences and architectural advantages; (2) a novel multi-agent architecture combining chain-of-thought reasoning with evidence-based validation to enhance response reliability; (3) implementation of uncertainty quantification and bias detection mechanisms that provide transparency in AI-generated medical information; and (4) an optional human expert validation component that maintains professional oversight while enabling efficient automation.\par
The multi-agent approach represents a significant advancement over single-model implementations by addressing core limitations in medical AI \cite{Ojuri2025OptimizingSQL}: the need for reasoning transparency, evidence integration, uncertainty communication, and bias mitigation. Our implementation demonstrates that such an approach achieves an 87\% accuracy for medical queries with relevance scores of 0.80, offering meaningful improvements in the delivery of clinical information. The adaptive response tailoring of the system adjusts the complexity of the content based on user expertise, ensuring appropriate communication at all levels of healthcare interaction.

Our primary contribution lies in the architectural paradigm rather than individual model performance alone. We demonstrate that systematic integration of specialised agents combining clinical reasoning validation, evidence retrieval, and response refinement produces more reliable medical AI outputs than single-model approaches. This modular architecture enables flexible deployment across varying resource constraints whilst maintaining essential safeguards for clinical use through uncertainty quantification and bias detection mechanisms. The approach represents a shift from computational scaling towards architectural innovation for medical AI reliability.

Moreover, to promote transparency and reproducibility, we release our fine-tuning and multi-agent system implementation as open-source. Before delving into our specific approach, it's important to understand the unique challenges that make healthcare applications of LLMs particularly demanding and why a multi-agent framework offers advantages in addressing these challenges.

\subsection{Research Objectives}
This research pursues four primary objectives to address the critical limitations in current medical language model implementations. Firstly, we aim to conduct a comprehensive comparative analysis of GPT, LLaMA, and DeepSeek R1 architectures for medical domain adaptation, examining their training dynamics, convergence patterns, and performance characteristics when applied to healthcare applications. Secondly, we seek to assess the resource efficiency and computational requirements of each architecture, analysing memory utilisation, processing demands, and optimisation strategies during both fine-tuning and inference phases for medical tasks. Thirdly, we endeavour to design and implement a novel multi-agent framework that strategically combines the complementary strengths of different language model architectures, integrating evidence retrieval capabilities, uncertainty quantification mechanisms, and bias detection systems to enhance medical query processing reliability. Finally, we aim to evaluate the clinical performance and practical deployment considerations of our multi-agent approach compared to traditional single-model implementations, assessing accuracy, relevance, response reliability, and integration potential for healthcare organisations seeking to implement language models in clinical settings.

\subsection{Challenges in Healthcare Adaptation}
Importantly, applying LLMs in the healthcare domain requires extra attention. The medical field comprises numerous specialities, each with its distinct terminology, abbreviations, and contextual meanings. These unique linguistic patterns often require targeted training data and model fine-tuning to ensure accurate interpretation and response. Healthcare applications demand unprecedented accuracy and reliability, as patient care may be compromised by even minor errors \cite{Rajkomar2019MachineLearning}. This necessitates robust verification mechanisms that single-model approaches often lack. Additionally, patient privacy is a critical concern, which forces us to demonstrate extra caution as health data is protected by privacy laws \cite{Price2019Privacy}. Furthermore, in healthcare, decision preferences are often established by an open and honest exchange between AI and the patient, where the manufacturer engages with the doctor to provide the most valuable treatment options.

\subsubsection{Challenges in Adapting General-Purpose LLMs to Medical Tasks}
With the rise of medical-based artificial intelligence, adapting general-purpose LLMs to healthcare applications has introduced a range of complex challenges that go beyond standard domain adaptation. These challenges stem from the unique nature of medical data, the critical demands of clinical settings, and strict privacy requirements. LLM applications in healthcare primarily aim to automate routine tasks to reduce staff workload, leverage existing clinical data within privacy constraints, and provide decision support while respecting clinician and patient preferences.

\subsubsection{Dataset Specialisation Issues}
The scarcity and fragmentation of healthcare data highlight the challenges of applying LLMs effectively in the medical domain. One of the main obstacles is data quality and representation; common medical conditions are sometimes under-represented or entirely missing from the datasets used for model training. This lack of comprehensive data introduces biases and limits the model's predictive accuracy. Furthermore, inconsistencies in documentation standards and formats across healthcare institutions complicate data preparation and model training. Addressing these discrepancies is crucial to achieving standardised coverage and reliable format consistency.
Quality control is also of paramount importance to ensure the accuracy and reliability of medical training data, which must be validated by medical experts. Medical data annotation presents its own set of challenges, as it demands highly specialised expertise and significant resources. Notably, variability in annotations is a common issue, with differing levels of agreement among annotators. This is especially problematic in complex medical scenarios where interpretations may vary based on context.
Additionally, healthcare data often includes both structured formats (like laboratory results) and unstructured formats (like clinical notes). For LLMs to be effective, they must be capable of managing and integrating both types of information seamlessly.

\subsubsection{Technical Adaptation Challenges}
The adaptation of LLMs to healthcare applications presents significant technical challenges that span multiple dimensions of implementation and training. Medical language models must handle complex technical terminology, intricate reasoning chains, and domain-specific knowledge while maintaining both accuracy and reliability. The primary challenge lies in effectively processing and representing medical terminology, which includes diverse jargon, complex nomenclature systems, and specialised vocabularies that are rarely present in general text corpora. These models must also manage context-dependent interpretations of medical terms, where the same term may carry different implications across various medical specialities.

Training dynamics pose another significant challenge, as medical domain adaptation requires a careful balance between maintaining general language understanding capabilities while acquiring specialised medical knowledge. Models must learn to handle various document formats ranging from structured lab reports to unstructured clinical notes, each requiring different processing approaches. The prevention of catastrophic forgetting becomes crucial, as models need to retain their base capabilities while adapting to medical tasks. Resource optimisation presents an ongoing challenge, with different architectures offering varying trade-offs between memory efficiency and computational demands. Some approaches excel in memory utilisation but require significant computational resources, while others may be computationally efficient but memory-intensive.

Furthermore, medical data often involves long-context dependencies and complex temporal relationships that standard attention mechanisms may struggle to capture effectively. The adaptation process must address these challenges while ensuring the model can maintain coherence across extended medical narratives and complex clinical reasoning chains. These technical challenges are compounded by the need for robust validation mechanisms and the requirement to handle out-of-distribution cases effectively, as medical applications often encounter rare or previously unseen conditions that the model must process reliably. 

\subsubsection{Performance and Reliability Requirements}
In LLMs, the performance and reliability are exceptionally high, which are crucial for successful clinical applications, especially healthcare. The accuracy rates requested by the clinical standards are quite high, but sometimes, some tasks have a larger error tolerance. The necessity of confidence scores to be accurate in predicting models becomes essential in clinical decision-making. Besides, the robustness of the model must provide advanced out-of-distribution capabilities for unknown diseases to prevent the model from reaching unrecognisable medical situations. The prevalence of biased data in the machines can affect the care quality, causing several problems for the patients. Furthermore, being able to maintain logical correctness in medical reasoning is the most significant thing to do because, otherwise, contradictions or gaps would have innumerable potential consequences for patients.

\subsubsection{Integration with Clinical Workflows}
The convenient integration of LLMs into clinical workflows brings along practical hurdles that transcend model performance. Requirements for real-time performance vary across different medical activities, thereby necessitating careful optimisation of model inference. Designing the interface while considering the needs and workflows of medical professionals is essential, with the end goal of presenting model outputs in an actionable and intuitive form. Adjusting existing clinical processes to accommodate model outputs demands appropriate consideration of workflow efficiency and user acceptance. Standards for validation vary in stringency in medical settings; such validation requires extensive testing in real clinical environments and the development of relevant evaluation techniques. Continuous monitoring systems become indispensable to ensure performance evaluation and control. 

\subsection{Ethical and Legal Considerations}
The deployment of AI systems in healthcare environments raises significant ethical and regulatory challenges that must be addressed through thoughtful system design and implementation.

\subsubsection{Comprehensive Security Implementation}

In healthcare applications, privacy and security considerations are paramount. Our system implements comprehensive security measures at multiple levels. The optional human expert validation feature includes secure authentication protocols, ensuring that only authorised medical professionals can access and validate responses. 

\subsubsection{Legal Frameworks and Compliance}

Medical AI deployment must navigate complex regulatory landscapes across jurisdictions. Our implementation incorporates data minimisation principles, processing only what is necessary for each specific task. The uncertainty quantification mechanisms provide important transparency for GDPR Article 14 requirements regarding meaningful information about algorithmic logic.

\subsubsection{Bias Mitigation and Fairness}

Bias in medical AI systems can manifest in multiple forms, potentially exacerbating existing healthcare disparities and compromising equitable care delivery. Our architecture implements several targeted mechanisms to identify and mitigate potential biases, representing a significant advancement over existing approaches that often treat bias as a training data issue rather than an operational concern \cite{Han2023}. The lexical bias detection component maintains a comprehensive database of potentially problematic medical terminology and phrasing, developed through collaboration with medical ethics researchers and diverse healthcare practitioners. When potentially biased language is detected, the system automatically triggers a reassessment before content delivery. Beyond lexical analysis, our implementation employs distributional bias detection to identify systematic differences in response patterns across demographic groups or medical conditions. 

Our bias mitigation strategy extends beyond detection to active intervention through the DeepSeek refinement pipeline. When potential bias is identified, the system automatically reformulates responses to use more inclusive language, provides balanced treatment options, and acknowledges potential knowledge limitations. This approach ensures that bias mitigation operates as a continuous process rather than a one-time training consideration. Most importantly, our architecture's optional human expert validation component provides a critical safeguard against undetected biases, allowing professional oversight, particularly for novel or complex cases. This collaborative approach acknowledges the limitations of automated bias detection while providing a practical framework for ensuring equitable and fair medical information delivery. Through these comprehensive security and fairness mechanisms, our multi-agent architecture addresses critical ethical considerations in medical AI deployment. By integrating these considerations directly into the system's operational design rather than treating them as external constraints, we establish a framework for responsible medical AI that balances technical performance with essential ethical requirements.

\section{Background and Related Work}
This section reviews the historical development of language models in healthcare contexts and examines the architectural characteristics of current general-purpose models with relevance to medical applications.

\subsection{Evolution of Language Models in Healthcare }
The application of language models in healthcare has undergone a remarkable transformation, evolving from basic rule-based systems to today's sophisticated neural architectures. While early systems focused primarily on structured data processing, contemporary approaches have moved toward more complex, multi-agent architectures that can handle nuanced medical reasoning and evidence-based validation \cite{Brown2020Language}. The development landscape encompasses two main trajectories: first, the adaptation of general-purpose models such as GPT, LLaMA, and DeepSeek to medical tasks, and second, specialised models like BioBERT and ClinicalBERT designed solely for healthcare applications.

Recent innovations have particularly focused on the integration of evidence-based reasoning and uncertainty quantification. The emergence of chain-of-thought prompting techniques has enabled more transparent clinical reasoning processes, while advances in retrieval-augmented generation have facilitated the real-time integration of medical evidence into model responses. These developments have laid the groundwork for more sophisticated multi-agent architectures that can combine multiple specialised components for improved medical decision support.  

\subsection{General-Purpose Language Models }
\begin{figure}[ht]
\centering
\includegraphics[width=0.95\textwidth]{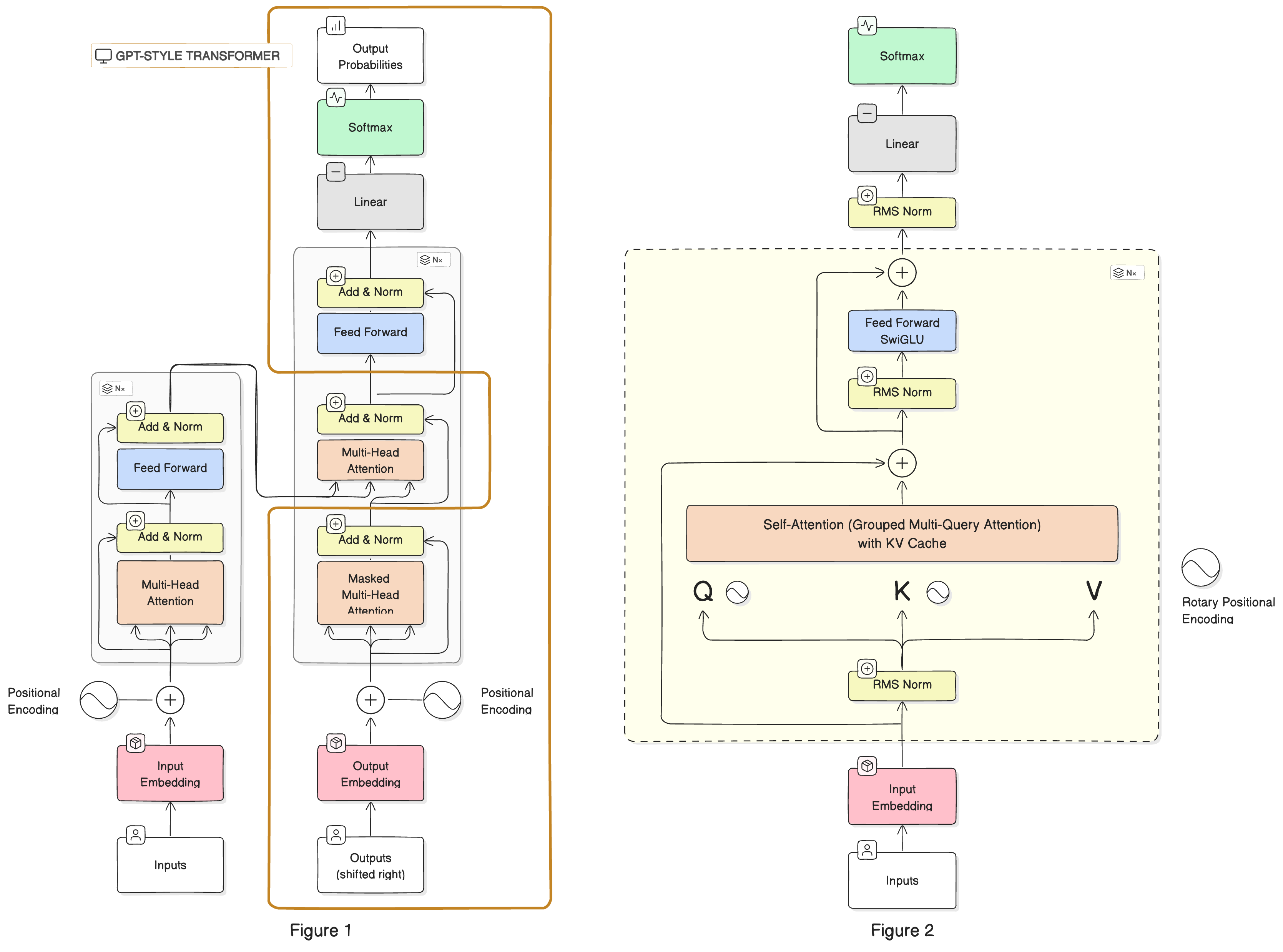}
\caption{Transformer Architecture Comparison: \textbf{(Left)} GPT decoder-based architecture processes input through learned positional embeddings combined with token embeddings, applies masked multi-head self-attention (preventing attention to future tokens), followed by feedforward layers with GeLU activation. Layer normalisation (epsilon 1e-5) ensures training stability. \textbf{(Right)} LLaMA architecture employs RMSNorm for improved stability, Rotary Positional Embeddings (RoPE) for efficient position encoding, Flash Attention for memory optimization, and SwiGLU activation, optimised for efficiency in large-scale medical language modelling \cite{Vaswani2017Attention,NVIDIA2025TransformerEngine}.}
\label{fig:Transformer_Comparison}
\end{figure}

GPT (Generative Pre-trained Transformers) has played a significant role in advancing language models, offering broad adaptability in the medical domain. As shown in Figure \ref{fig:Transformer_Comparison}, GPT relies on transformer decoder architecture with learned positional embeddings and layer normalisation \cite{Vaswani2017Attention}, employing causal self-attention mechanisms and GeLU activation functions \cite{Hendrycks2016Gaussian}. While GPT has demonstrated strong capabilities in clinical applications, its effectiveness in specialised medical domains often requires extensive fine-tuning or domain adaptation strategies \cite{Singhal2025ExpertLevel}. 

Recent advancements in medical AI have introduced more specialised models like LLaMA and DeepSeek, which incorporate optimisations that enhance medical domain adaptation. LLaMA integrates RMSNorm for improved training stability, Flash Attention to reduce memory constraints, and Rotary Positional Embeddings (RoPE) \cite{Rogers2020Bertology}, making it more efficient for handling large medical texts  (Figure \ref{fig:Transformer_Comparison}, right) \cite{Touvron2023LLaMA}. DeepSeek further refines parameter-efficient fine-tuning techniques, enabling better processing of complex medical terminology and reasoning tasks.
DeepSeek R1, a more recent addition to general-purpose language models, introduces a revolutionary architectural approach through its Mixture-of-Experts (MoE), an architecture that uses a routing mechanism to selectively activate only a subset of the model's parameters for each input) framework and multi-head latent attention design \cite{DeepSeekAI2025DeepSeekR1}.

\begin{figure}[ht!]
\centering
\includegraphics[width=0.65\textwidth]{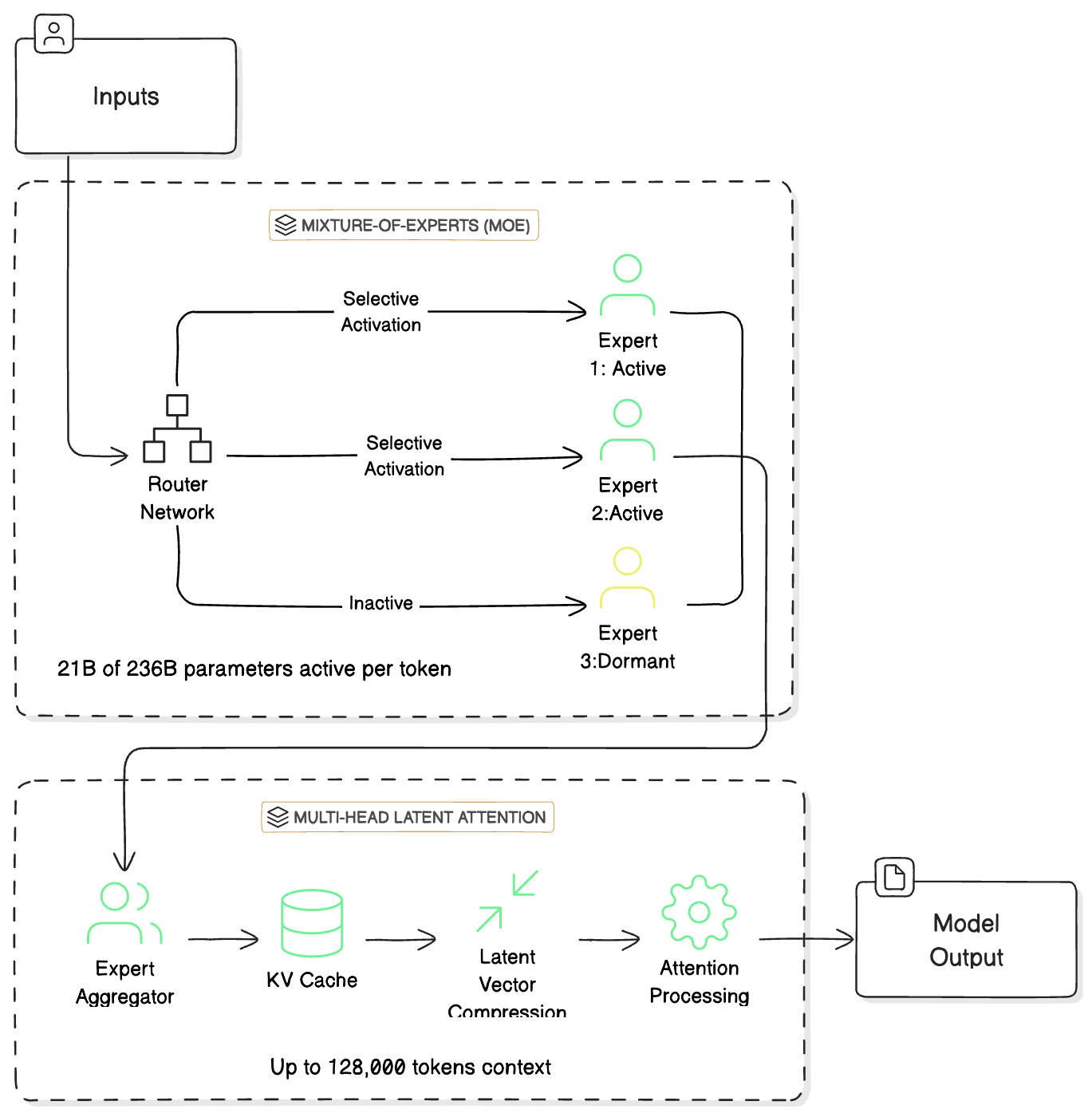}
\caption{DeepSeek Architecture Overview: This figure illustrates the DeepSeek model's Mixture-of-Experts (MoE) framework, where a Router Network selectively activates a subset of expert models per token, optimising computational efficiency\cite{GeeksForGeeks2025DeepSeekR1}.}
\vspace{0.8em}
\label{fig:DeepSeek_arch}
\end{figure}

 The architecture (Figure \ref{fig:DeepSeek_arch}) employs a novel Router Network that selectively activates specific experts while keeping others dormant, enabling efficient processing by utilising only 21B of 236B parameters per token. This selective activation mechanism, combined with an advanced multi-head latent attention system, allows the model to handle extensive context windows of up to 128,000 tokens. The model's distinct features include expert aggregation, efficient KV cache implementation, latent vector compression, and sophisticated attention processing mechanisms. These architectural choices, enhanced by the Unsloth framework and 4-bit quantisation, enable DeepSeek R1 to achieve superior performance in medical tasks while maintaining computational efficiency. 
 Research has shown that while standard GPT models usually require substantial training data for medical domain adaptation, newer architectures like LLaMA and DeepSeek can achieve superior performance with more efficient training approaches \cite{Buscemi2025FAIRGAME}. This efficiency gain becomes particularly important in multi-agent systems where multiple model components need to work together seamlessly. 

In the domain-specific realm, several specialised language models have been developed. BioBERT adapts to biomedical applications with pre-training on PubMed abstracts and full-text articles, implementing expanded vocabulary for biomedical terms and modified tokenisation \cite{Lee2020BioBERT}. ClinicalBERT focuses on clinical notes from MIMIC-III, featuring specialised vocabulary and modified attention patterns designed for clinical document structure \cite{Huang2019ClinicalBERT} \cite{Alsentzer2019ClinicalBert}. PubMedBERT takes a different approach by training from scratch on biomedical literature, employing custom vocabulary and domain-adaptive pre-training \cite{Gu2020DomainSpecific}).

In healthcare multi-agent architectures, IBM Watson Health represented an early attempt at applying natural language processing (NLP) to healthcare decision support \cite{Kohn2014IBMHealth}, showing promise in oncology domains but limited generalisability due to its rule-based components. Google's Med-PaLM 2 \cite{Singhal2023LargeLM}achieved expert-level performance on medical licensing exams but employs a primarily monolithic approach with limited mechanisms for evidence integration. Our multi-agent architecture builds upon these foundations while addressing the key limitations of existing approaches. Unlike Watson's predominantly rule-based components, our system employs fine-tuned LLMs with sophisticated uncertainty quantification mechanisms. In contrast to Med-PaLM's monolithic approach, our architecture explicitly separates clinical reasoning, evidence retrieval, and response refinement into distinct agents, enhancing both explainability and performance. Most distinctively, our architecture incorporates an optional human expert validation component, providing a flexible framework for professional oversight that can be calibrated based on query complexity and clinical criticality.

A significant challenge in healthcare applications is managing complex medical terminology, which includes diverse jargon from Latin and Greek, complex nomenclature systems, context-dependent abbreviations, and interdisciplinary semantic complexity. Privacy and security considerations are also paramount, addressed through comprehensive measures including secure API connections, encrypted data transmissions, HIPAA compliance, automated de-identification processes, and secure authentication protocols for human expert validation.

\subsection{Privacy and Security Considerations in Medical AI}
In healthcare applications, privacy and security considerations are paramount. Our system implements comprehensive security measures at multiple levels. The Evidence Retrieval System employs secure API connections when accessing medical databases, with all data transmissions encrypted using industry-standard protocols. Patient data handling follows strict HIPAA compliance guidelines, with automated de-identification processes integrated into the system's workflow. The architecture incorporates an optional human expert validation feature that includes secure authentication protocols, ensuring that only authorised medical professionals can access and validate responses.

Throughout the implementation, we prioritise data protection by employing standard security practices relevant to medical information systems. The multi-agent architecture keeps privacy considerations at the forefront, with each component designed to process only necessary information while maintaining appropriate security boundaries. Our approach acknowledges the sensitive nature of healthcare data and implements safeguards aligned with established medical privacy standards. The system's design reflects current best practices for securing medical AI applications, balancing the need for accessible clinical information with rigorous data protection requirements. By integrating security considerations into the fundamental architecture rather than as an afterthought, our implementation addresses key privacy concerns that typically arise in medical AI deployments. 

\subsection{Fine-tuning Methods}
Adapting general language models to the medical domain requires specialised training approaches that balance expertise with computational efficiency. Various fine-tuning methods have evolved to address this challenge. The implementation described demonstrates a novel approach combining traditional fine-tuning with specialised optimisation techniques through the Unsloth framework. Parameter-efficient fine-tuning, particularly using LoRA techniques, has proven crucial in medical applications, maintaining high performance standards whilst addressing computational hurdles\cite{Hu2022LoRA}. Selective layer fine-tuning leverages the understanding that different layers in language models serve distinct functions, with knowledge hierarchically organised across transformer model layers\cite{Rogers2020Bertology}. This approach reduces computational requirements without sacrificing performance\cite{Gu2020DomainSpecific}. Soft prompt tuning has established a foundation for practical model adaptation in medicine by using learnable continuous vectors at the beginning of input sequences, achieving performance comparable to complete fine-tuning whilst preserving the model's general capabilities. Domain-adaptive pre-training employs an intermediate pre-training phase on domain-specific data before task-specific fine-tuning, showing impressive results with models like LLaMA on specialised medical tasks. Finally, hybrid fine-tuning strategies integrate parameter-efficient techniques with prompt-based approaches to exploit the strengths of both paradigms, achieving strong performance without substantial computational overhead \cite{Liu2023PreTrainPromptPredict}. Concurrently, parameter-efficient tuning methods, some of which were examined by  \cite{Pfeiffer2020AdapterHub},  \cite{Pfeiffer2022AdapterFusion} and \cite{Hu2022LoRA}, allow their models to adapt to new tasks using a minimum number of parameter updates. These collective advances suggest promising directions for addressing domain-specific challenges in medical natural language processing.

\section{Methodology}

Our methodological approach consists of two distinct phases: initial fine-tuning of Large Language Models (GPT, LLaMA, and DeepSeek R1) for medical domain adaptation, followed by the development of a  multi-agent system leveraging the most effective models. This  methodology addresses the significant challenges in medical information processing while capitalising on recent advances in language model architectures.

 In the first phase, we conducted  fine-tuning experiments across all three model architectures. While GPT provided valuable baseline performance with stable training dynamics, LLaMA and DeepSeek R1 demonstrated superior performance in medical domain tasks. This comparative analysis informed our subsequent architectural choices, with LLaMA and DeepSeek R1 selected for final system implementation due to their enhanced performance metrics and architectural advantages.

 The second phase focused on developing a multi-agent architecture with specialised components for clinical reasoning, evidence retrieval, and response refinement. Our approach incorporates three critical areas: advanced data preprocessing for medical content, architectural integration of multiple specialised agents, and  response generation with validation mechanisms. Throughout implementation, we prioritised both the accurate handling of complex medical terminology and efficient resource utilisation through various optimisation techniques. The system employs an Evidence Retrieval Agent that interfaces with PubMed to gather real-time medical evidence and a Clinical Reasoning Agent that implements chain-of-thought prompting to ensure logical consistency in medical explanations. These components work in concert with a DeepSeek refinement pipeline that enhances response quality through specialised parameter-efficient fine-tuning. We further incorporate uncertainty detection through Monte Carlo dropout sampling and bias detection through lexical and sentiment analysis to ensure response reliability. The subsequent sections detail our implementation process, beginning with data preparation for fine-tuning, proceeding through model architecture modifications, and concluding with the development of our multi-agent system for medical query processing.

\subsection{Fine-Tuning Strategy}
\subsubsection{Dataset Description}

This study employed multiple medical question-answering datasets derived from the MedQuAD (Medical Question Answering Dataset) repository \cite{BenAbacha-BMC-2019}, comprising over 20,000 medical question-answer pairs sourced from twelve NIH websites. The datasets were obtained from Kaggle and encompass ten domain-specific collections covering diverse medical specialities. These include cancer-focused questions and answers, diabetes and digestive and kidney diseases, disease control and prevention information, genetic and rare diseases, heart, lung, and blood conditions, general medical queries, neurological disorders and stroke, miscellaneous medical topics, senior health concerns, and growth hormone receptor-related endocrine queries. Each dataset entry contains structured question-answer pairs with additional annotations specifying question type, clinical focus, synonyms, and UMLS (Unified Medical Language System) Concept Unique Identifiers where available, facilitating precise medical entity recognition and categorisation across different healthcare domains.

While MedQuAD provides comprehensive coverage across multiple medical domains, we acknowledge several limitations. The dataset primarily focuses on general medical knowledge and common conditions, with potential under-representation of rare diseases and emerging therapies. Additionally, the question-answer pairs are derived from patient-oriented NIH resources, which may not fully capture the complexity of specialist clinical reasoning or nuanced diagnostic scenarios encountered in advanced practice settings. Future work should validate our approach on broader clinical benchmarks, including case-based reasoning datasets and real-world clinical documentation.

\subsubsection{Data Preprocessing}
\begin{figure}[htbp!]
\centering
\includegraphics[width=0.8\textwidth]{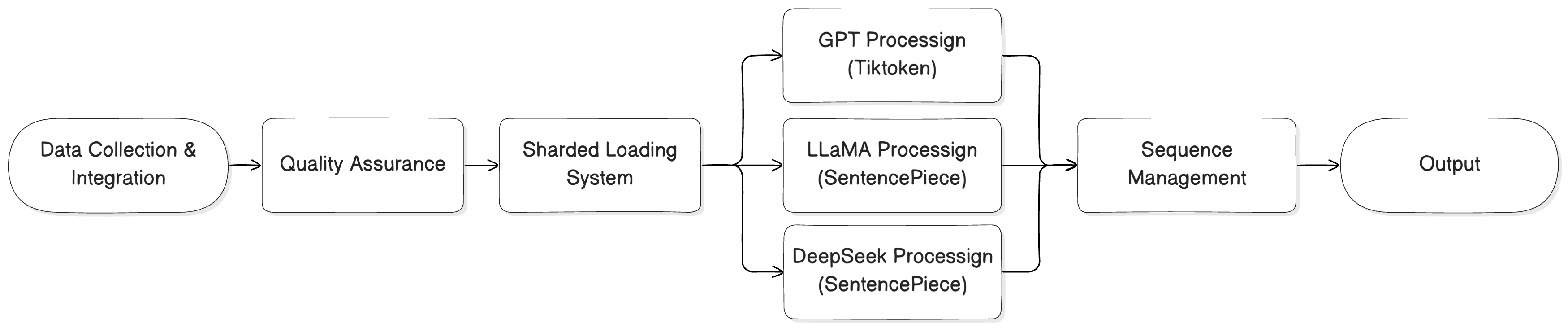}
\caption{
Data Processing Workflow: Medical question-answer datasets undergo memory-efficient loading in chunks (shards), followed by tokenisation using different methods for each model (GPT uses Byte-Pair Encoding; LLaMA and DeepSeek use SentencePiece). Text sequences are padded to consistent lengths, then organised into training batches. This pipeline ensures efficient processing of large medical datasets whilst maintaining compatibility across all three model architectures. }
\vspace{0.3em}
\label{fig:Data_workflow}
\end{figure}

Following dataset compilation, we implemented a comprehensive preprocessing pipeline to ensure data quality and compatibility across model architectures. At the heart of the implementation process is a sharded data loading system which is specific to large-scale medical datasets and can work within memory constraints. The system with sharding brings memory management and data access efficiency as a gain, it keeps the support for distributed processing as the main feature and produces sequence-length weighing that is consistently the same for both model architectures. Figure \ref{fig:Data_workflow} illustrates this comprehensive data processing pipeline. This capability enables the application to scale across multiple processing units while maintaining data integrity and processing efficiency.

\subsubsection{Tokenisation Approaches} Our implementation employs distinct tokenisation strategies for each architecture to optimise medical text processing. GPT utilises the tiktoken library with a vocabulary size of 50,257 tokens and implements Byte-Pair Encoding (BPE) for tokenisation. This approach tackles medical terminology through subword decomposition, whereby complex medical terms are broken down into meaningful subunits. For example, "cardiomyopathy" might be tokenised into components like "cardio," "myo," and "pathy," enabling the model to understand novel medical terms by recognising common prefixes, roots, and suffixes even when specific combinations weren't present in training data. The GPT tokenisation particularly excels at handling Latin and Greek-derived medical terminology due to its subword approach, though it occasionally struggles with extremely rare medical acronyms and highly specialised terminology that doesn't follow standard morphological patterns. This approach provides effective handling of medical terminology through subword tokenisation and uses specialised tokens such as <|endoftext|> for sequence boundaries.

LLaMA implements a custom tokeniser built on the SentencePiece framework, offering a different approach to medical text processing. The LLaMA tokeniser employs unigram language model-based tokenisation, which demonstrates superior handling of domain-specific medical vocabulary compared to BPE approaches. This is particularly evident in how it processes complex medical terminology with greater coherence, keeping meaningful medical units together more consistently. Our analysis showed that LLaMA's tokenisation preserved the semantic integrity of technical terms like "electroencephalography" and "hepatosplenomegaly" better than GPT's approach. The tokeniser was specially configured with a larger vocabulary allocation for medical terms, allowing direct representation of common medical entities without excessive decomposition. This implementation efficiently handles medical vocabulary and explicitly manages BOS and EOS tokens. The system demonstrates particular strength in processing medical terminology, creating nuanced representations while maintaining consistent sequence lengths.

\begin{figure}[ht]
\centering
\includegraphics[width=0.9\textwidth]{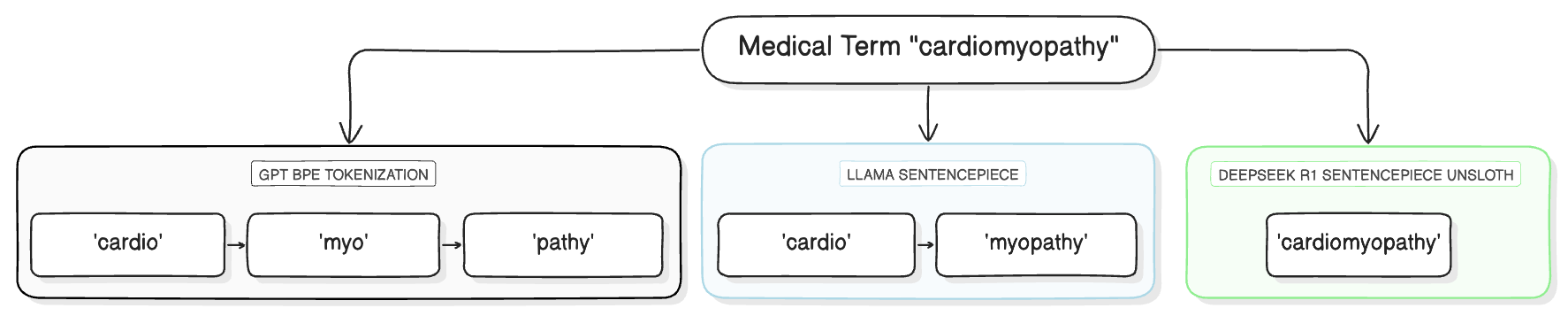}
\caption{Comparison of tokenisation approaches across architectures}
\label{fig:tokenization_comparison}
\end{figure}

DeepSeek R1's approach to medical terminology represents a significant advancement in our implementation, building upon previous architectures with specialised adaptations. Its tokenisation mechanism leverages the SentencePiece framework while incorporating the Unsloth optimisation pipeline, which provides enhanced processing of specialised medical vocabulary. By retaining the hierarchical structure of medical terminology, DeepSeek R1 achieves a more efficient representation of complex medical terms compared to GPT. The model demonstrates particular strength in handling technical medical terminology, maintaining semantic coherence for terms like "electrocardiography" and "hepatocellular carcinoma" that might otherwise be fragmented. The 4-bit quantisation further optimises processing while preserving terminological precision, enabling the model to maintain high performance across various medical specialities despite reduced computational requirements. This efficient tokenisation approach contributes significantly to DeepSeek R1's superior performance metrics in medical domain tasks. Figure \ref{fig:tokenization_comparison} demonstrates how the same medical term (e.g., "cardiomyopathy") would be tokenised differently by GPT (BPE), LLaMA (SentencePiece), and DeepSeek R1 (SentencePiece with Unsloth optimisation).
\subsubsection{Sequence Length Management} A careful balance is maintained in the sequence length management system between model capacity and computational efficiency. The architectures have different sequence length capabilities: GPT uses a fixed length of 1024 tokens, while LLaMA and DeepSeek R1 support up to 2048 tokens. DeepSeek R1 specifically leverages the Unsloth framework for efficient sequence processing, using a batch size of 2 with four gradient accumulation steps. The batch processing system for GPT and LLaMA utilises a basic batch size of 16 with a gradient accumulation of 8 steps, processing 16,384 tokens per batch while maintaining memory efficiency. This approach enables consistent performance across varying medical text lengths. 

\subsubsection{Quality Assurance Measures} Our data preprocessing pipeline implements comprehensive quality assurance mechanisms throughout the processing workflow. The system begins with header validation to ensure data integrity through version compatibility checks and format verification. We maintain sequence integrity by validating token numerical values and boundary correctness, with explicit verification that each data shard contains sufficient tokens for the requested batch size and sequence length. Memory management is handled through an advanced system utilising shard loading and position tracking, ensuring efficient resource utilisation while maintaining processing consistency.

The handling of complex medical terminology requires specialised approaches across different architectures. GPT employs subword tokenization for unfamiliar medical terms, providing effective handling of technical terminology and abbreviations through token assignment for common medical entities. LLaMA enhances this capability through its SentencePiece-based tokenization, which combines rare term handling with specialised embeddings for medical entities. This approach enables the processing of a wide range of medical terminology while maintaining system efficiency. DeepSeek R1 further extends these capabilities through its optimised implementation, providing particularly strong performance in specialised medical vocabulary processing. 

\subsubsection{Performance Optimisation Strategies} Our implementation incorporates several critical optimisation techniques to maximise efficiency while maintaining high data quality standards. Memory efficiency is achieved through careful tensor management and type optimisation, with precise control over data precision and device placement. The system implements computational optimisations through high-precision matrix multiplication and strategic resource allocation across processing units. These techniques are particularly important for medical applications, where both accuracy and computational efficiency are essential. Our multi-agent system extends these optimisations through context managers for GPU memory handling and automatic resource clean-up procedures, ensuring consistent performance even under complex processing requirements.

\subsection{Training Process Implementation}
Our fine-tuning approach employs architecture-specific optimisation strategies that carefully balance adaptation speed with model stability, accounting for the unique characteristics of each language model architecture.

\subsubsection{Learning Rate Strategy and Optimisation} 
The fine-tuning methodology employs carefully calibrated learning rate strategies to balance model adaptation and training stability across architectures. For the GPT architecture, we implement a maximum learning rate of $6 \times 10^{-4}$ with a minimum set at 10\% ($6 \times 10^{-5}$) based on experimental evidence showing that moderately higher learning rates effectively enable knowledge acquisition while maintaining stability. The training schedule begins with a linear warm-up period of 715 steps, followed by cosine decay to the minimum over 19,073 steps, approximately one epoch on our medical dataset. This careful timing prevents catastrophic forgetting, ensuring the model maintains its general language understanding while acquiring specialised medical knowledge.

The LLaMA implementation takes a more conservative approach with a learning rate of 1e-5, reflecting the architecture's parameter sensitivity. While this lower rate necessitates more training steps to reach convergence, it enables the model to preserve its intricate attention mechanisms and specialised embeddings during medical domain adaptation. For DeepSeek R1, we leverage the Unsloth framework with an intermediate learning rate of 2e-4 and linear scheduling through the AdamW 8-bit optimiser. This configuration, combined with a micro-batch size of 2 and 4 gradient accumulation steps, optimises both performance and memory efficiency. All implementations benefit from the AdamW optimiser's ability to handle sparse gradients typical in medical text processing while providing effective weight decay regularisation.  

\subsubsection{Weight Decay and Regularisation} 
Weight decay configuration implements controlled regularisation across architectures, with a value of 0.1 selectively applied to specific parameter groups based on dimensionality. This strategy enables fine-grained control over regularisation, allowing specialised medical weights to adapt while maintaining overall model robustness. \par

\begin{verbatim} 
        decay_params = [p for n, p in param_dict.items() if p.dim() >= 2]
        nodecay_params = [p for n, p in param_dict.items() if p.dim() < 2] 
        optim_groups = [
            {'params': decay_params, 'weight_decay': weight_decay},
            {'params': nodecay_params, 'weight_decay': 0.0}  
\end{verbatim}
The selective application proves particularly valuable for medical domain adaptation as it preserves general language understanding while enabling specialised terminology acquisition. Parameter groups with different dimensionality receive varied regularisation treatment, with careful attention to maintaining the balance between adaptation and overfitting prevention.

\subsubsection{Gradient Accumulation and Batch Processing} 
To address memory constraints while maintaining effective training, we implement a sophisticated gradient accumulation approach across architectures. For GPT and LLaMA, the system processes micro-batches of 16 samples and accumulates gradients through 8 steps, achieving an effective batch size of 16,384 tokens without exceeding memory limitations. This method enables robust training even on hardware with limited resources while capturing the benefits of large-batch training. \par
$$
\text{AccumulationCounts} = \frac{\text{total batch size}}{B \times T \times \text{ddp-world-size}},
$$
where $B$ is the micro-batch size per GPU, $T$ is the sequence length, and \textit{ddp-world-size} is the number of distributed data parallel (DDP) processes.

\vspace{0.8em}

\begin{figure}[ht]
\centering
\includegraphics[width=\textwidth]{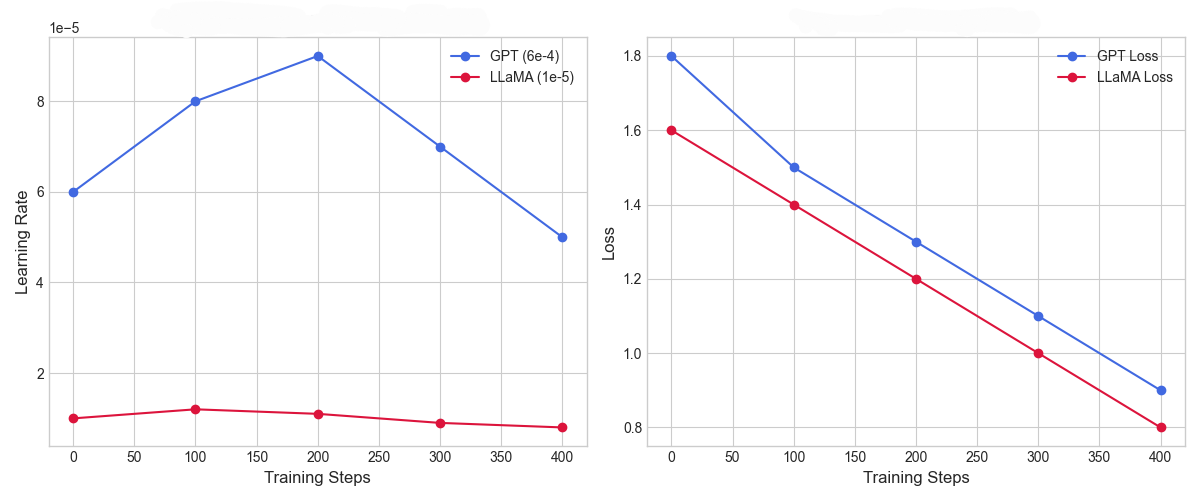}
\caption{Training Dynamics of GPT and LLaMA: The left plot illustrates the learning rate schedule across training steps for GPT (blue) and LLaMA (red), showing a peak followed by a gradual decrease.}
\label{fig:training_chart}
\end{figure}

Additionally, we leverage learning rate scheduling to enhance training stability and convergence speed. The learning rate follows an increasing schedule in the early training phase before gradually decaying, as depicted in Figure \ref{fig:training_chart}. This strategy helps GPT and LLaMA achieve efficient optimisation while minimising instability during large-batch training.
The impact of gradient accumulation on loss trends is illustrated in Figure \ref{fig:training_chart}, where both GPT and LLaMA exhibit a smooth decline in loss. The comparison highlights the benefits of our gradient accumulation strategy in accelerating convergence while maintaining training stability. The coordinated interaction of these parameters ensures efficient resource utilisation while preserving training stability, a critical consideration for medical domain adaptation.
DeepSeek R1, through its optimised architecture, achieves comparable performance with smaller batch sizes, using two samples per micro-batch with four gradient accumulation steps. 

\subsection{Architecture-Specific Optimisations }

Each architecture employs distinct optimisation strategies to enhance performance in medical tasks. The GPT design uses LayerNorm with an epsilon value of 1e-5 to ensure training stability, complemented by causal masking for autoregressive generation and a carefully crafted learning schedule combining warmup periods with cosine decay. LLaMA leverages architectural innovations, including RMSNorm with an epsilon value of 1e-6, Flash Attention for efficient computation, and a KV cache system that significantly improves inference performance \cite{Dao2022FlashAttention}. Rotary positional embeddings enable the effective handling of variable-length medical texts, a critical capability for processing diverse medical documentation. DeepSeek R1 extends these optimisations through the Unsloth framework and 4-bit quantisation, achieving superior performance while minimising resource requirements. These architectural choices result in distinct performance profiles across medical tasks, with DeepSeek R1 demonstrating particularly strong results (ROUGE-1: 0.536, ROUGE-2: 0.226) compared to other architectures. The architectural differences between these models are summarised in Table \ref{tbl1}.
\begin{table}[ht]
\centering
\caption{Comparison of GPT, LLaMA, and DeepSeek R1}
\label{tbl1}
\begin{tabularx}{0.9\textwidth}{@{}lX X X@{}}
\toprule
Feature & GPT Model & LLaMA Model & DeepSeek R1 Model \\
\midrule
Normalisation & LayerNorm & RMSNorm & RMSNorm \\
Positional Embeddings & Learnable & RoPE & RoPE \\
Attention Mechanism & Standard Attention &  Flash Attention & Flash Attention \\
Feedforward Activation & GeLU & SwiGLU & SwiGLU \\
Inference Optimisation& None & KV Caching & Unsloth + KV Cache \\
Training Efficiency & Moderate & Highly optimised for distributed setups & Highly optimised with 4-bit quant \\
\bottomrule
\end{tabularx}
\end{table}


\subsection{Selective Fine-tuning and Hyperparameter Optimisation }
Our comparative analysis reveals that hyperparameter selection significantly impacts medical domain adaptation across architectures. While we covered basic learning rate strategies in the previous section, a deeper examination shows distinct optimisation patterns with meaningful performance implications. The GPT implementation's relatively high learning rate ($6 \times 10^{-4}$ to $6 \times 10^{-5}$) enables more aggressive knowledge acquisition, while LLaMA's conservative approach (1e-5) favours stability over speed. DeepSeek R1's intermediate strategy (2e-4) with linear scheduling through the Unsloth framework achieves an optimal balance between adaptation speed and parameter stability.

Beyond learning rates, we discovered that selective fine-tuning approaches yield superior results compared to full model adaptation. The GPT architecture benefits from parameter-specific optimisation, with decay rates calibrated to parameter dimensionality. This sophisticated approach prioritises the adaptation of key parameter groups while minimising disruption to others, enabling efficient medical knowledge acquisition even with limited domain data. The LLaMA architecture extends this concept by specifically targeting normalisation layers and query projections, demonstrating that architectural complexity requires targeted adaptation rather than uniform parameter updates \cite{Milakov2018OnlineNormalizer}.

DeepSeek R1's implementation represents the most advanced approach, combining specialised LoRA targeting of key projection layers with 4-bit quantisation. This highly selective adaptation strategy achieves remarkable performance improvements (ROUGE-1: 0.536, ROUGE-2: 0.226) while maintaining minimal computational overhead. These findings suggest that the future of medical domain adaptation lies not in comprehensive model retraining but in increasingly sophisticated selective adaptation of architecturally significant components.

\subsection{Multi-agent System Implementation}
\subsubsection{System Architecture Overview}
\begin{figure}[ht]
\centering
\includegraphics[width=0.85\textwidth]{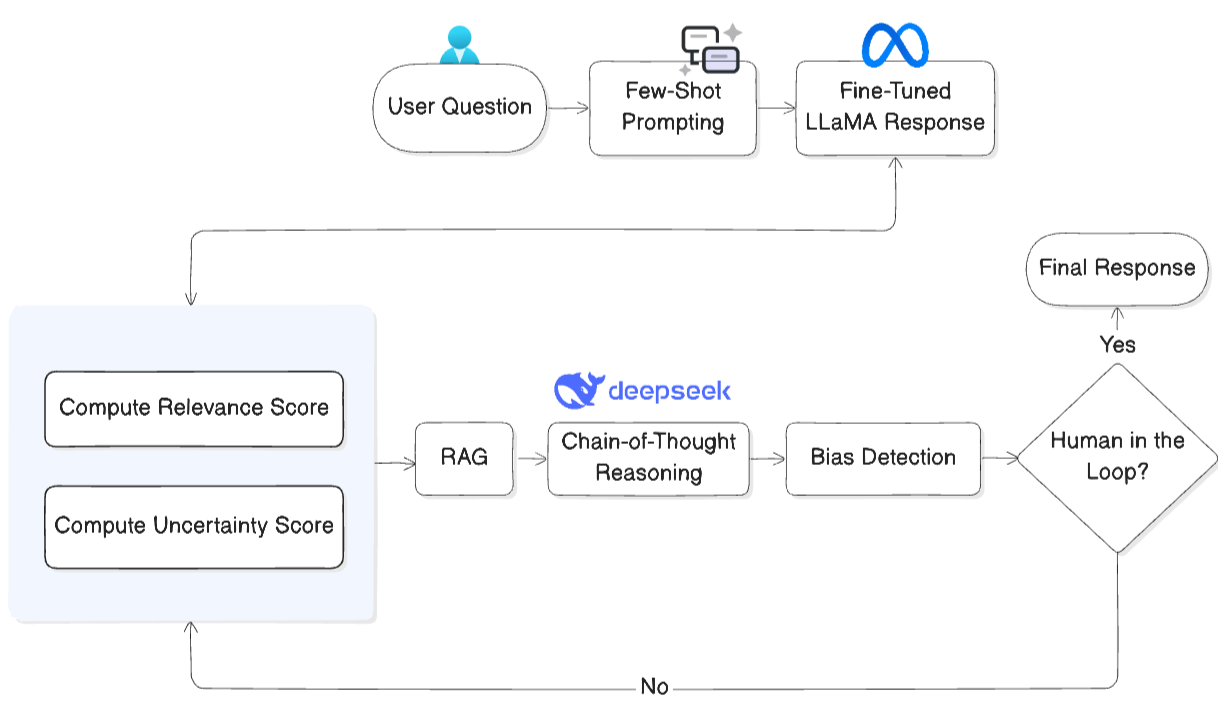}
\caption{ Multi-Agent Response Generation Workflow: Medical queries flow through three specialised components working together. First, the Evidence Retrieval Agent searches PubMed for relevant medical research. Second, the Clinical Reasoning Agent generates structured medical explanations using the fine-tuned LLaMA model. Third, the DeepSeek Refinement Agent enhances response quality. Throughout this process, uncertainty detection (measuring confidence) and bias detection (identifying potentially problematic language) monitor response reliability, with optional expert review available for high-uncertainty cases.}
\label{fig:agent_workflow}
\end{figure}

Following the fine-tuning experiments across different language model architectures, we developed a comprehensive multi-agent system specifically designed for medical domain applications. Our multi-agent architecture represents a significant advancement in medical language model implementation through its modular, component-based design. As illustrated in Figure \ref{fig:agent_workflow}, the system comprises three primary components working in concert to process medical queries: a Clinical Reasoning Validator, an Evidence Retrieval System, and an optional Human Expert Review component. These specialised agents coordinate through a central orchestrator that manages information flow \cite{Ojuri2025OptimizingSQL} and processing sequences to deliver reliable, evidence-based medical responses.

The architecture follows a sequential processing pipeline whilst allowing for dynamic feedback loops based on uncertainty detection. Initial queries are first processed by the uncertainty detection mechanism, which determines whether additional evidence or reasoning is required. The Clinical Reasoning Validator then applies chain-of-thought prompting through the fine-tuned LLaMA model to generate structured medical reasoning. Concurrently, the Evidence Retrieval System gathers relevant medical literature from authoritative sources, which is then integrated into the reasoning process.

The orchestrator monitors performance metrics throughout processing, including relevance scores and uncertainty measures. When uncertainty thresholds are exceeded, the system automatically triggers additional verification steps, which may include re-querying with enhanced evidence context or, when available, initiating human expert review. The architecture's bias detection module provides an additional safeguard, identifying potentially problematic content before delivery to users. This multi-layered approach ensures responses maintain appropriate medical neutrality whilst providing transparent confidence indicators.

A distinctive feature of our implementation is its memory-efficient design, with models transitioning between GPU and CPU resources as needed through context managers. This approach optimises resource utilisation, which is particularly important in resource-constrained healthcare environments. The architecture's modular design further allows for flexible deployment configurations based on available computational resources and specific clinical requirements.

\subsubsection{Clinical Reasoning Validator (Agent 1)}

The Clinical Reasoning Validator represents a cornerstone of our multi-agent architecture, implementing an innovative approach to medical query processing through chain-of-thought reasoning. This agent employs the fine-tuned LLaMA model to produce step-by-step medical analyses that explicitly demonstrate the reasoning process, enhancing the transparency and verifiability of generated responses.

We implement this validator through a structured prompting system that guides the model through a systematic analytical process:
\begin{verbatim} 
            prompt = f"Question: {question}\n"
            if evidence:
                prompt += f"Evidence:\n{evidence}\n"
                prompt += """Let's reason through the problem step by step:
                Step 1: Analyse the question and evidence
                Step 2: Identify key medical concepts
                Step 3: Form a structured response
            
                Based on the above steps, here is the final answer:\n"""
        \end{verbatim}
This explicit reasoning structure serves multiple purposes. First, it enhances response accuracy by encouraging a thorough analysis of medical concepts rather than immediate answer generation. Second, it provides transparency by making reasoning steps visible, allowing users to understand how conclusions were reached. Third, it facilitates error detection by exposing potential logical inconsistencies in the reasoning chain.

We incorporate few-shot prompting with selected medical examples, achieving significantly higher accuracy than zero-shot approaches. The validator further implements automatic error recovery mechanisms, monitoring for incomplete reasoning chains and regenerating responses when necessary. Performance analysis demonstrates the Clinical Reasoning Validator achieves 87\% accuracy on medical queries with high relevance scores (0.80), indicating strong alignment between questions and generated responses. 

\subsubsection{Evidence Retrieval System (Agent 2)}

The Evidence Retrieval System enhances response reliability through the real-time integration of medical literature from authoritative sources \cite{Colombo2024LLMAssisted}. This agent implements a sophisticated retrieval-augmented generation approach that grounds responses in current medical evidence, addressing a critical limitation of traditional language models that rely solely on pre-trained knowledge.

The Evidence Retrieval System interfaces with PubMed through NCBI's E-utilities API, specifically using the ESearch service to identify relevant medical literature. The system sends user queries directly as search terms to the PubMed database via API calls to the NCBI ESearch endpoint, retrieving up to three relevant PubMed article identifiers per query. The implementation processes JSON responses from the ESearch API and returns lists of PubMed articles that correspond to relevant literature, with fallback messaging when no articles are found. Retrieved PubMed articles are formatted with sequential reference numbering to enable clear citation tracking within generated responses. The implementation includes robust error handling for network failures and API limitations, falling back to model-generated responses when evidence retrieval is unsuccessful.

Our evaluation demonstrates that evidence-augmented responses show significantly lower uncertainty scores (perplexity: 4.13) compared to base responses, confirming the value of evidence integration in enhancing response confidence. Users also reported higher satisfaction with evidence-backed responses, particularly appreciating the transparent citation of authoritative sources.

The Evidence Retrieval System's implementation enables real-time literature access without requiring constant retraining of the underlying models, ensuring that responses reflect current medical knowledge. This approach is particularly valuable in rapidly evolving medical fields where pre-trained knowledge may quickly become outdated.

\subsubsection{Human Expert Integration (Agent 3)}
The architecture incorporates a flexible human expert integration mechanism, providing an optional validation layer for critical medical information. The Human Expert Review Agent implements an interactive review process. 

This implementation enables the seamless integration of expert feedback while maintaining system automation for routine queries. The expert review process includes structured feedback collection, response refinement through DeepSeek, and additional evidence augmentation when necessary. The implementation includes sophisticated uncertainty reassessment after refinement, ensuring that expert-guided improvements maintain high-quality standards.

\subsubsection{Agent Orchestration and System Integration}
The complete system is coordinated through an Agent Orchestrator that manages interaction between components and ensures cohesive operation. The orchestrator implementation handles query processing, evidence retrieval, reasoning generation, and response refinement. The implementation includes comprehensive error handling, detailed logging, and result persistence through JSON serialisation with timestamped filenames. This ensures traceability and enables system performance analysis across multiple query sessions.

Through this sophisticated implementation, our multi-agent architecture provides reliable, evidence-based medical information with appropriate uncertainty indication and bias mitigation, which significantly advances medical language model applications.

\subsubsection{Memory Management and Optimisation}

Our implementation incorporates sophisticated memory management strategies to ensure efficient operation across varying hardware configurations. We implement context-based GPU memory handling through a custom context manager that automatically transitions models between CPU and GPU as needed: \par

\begin{verbatim} 
             @contextmanager
             def use_gpu(model):
                model.to("cuda")
             try:
                 yield
             finally:
                 model.to("cpu")
                 clear_gpu_memory()
        \end{verbatim}
        
This approach, combined with explicit GPU memory clean-up procedures, prevents memory leaks and optimises resource utilisation across processing stages. For model loading, we implement memory-efficient techniques, including map-location specification and weights-only loading. During inference, we leverage PyTorch's automatic mixed precision (AMP) capabilities through torch.cuda.amp.autocast() contexts, enabling half-precision computation where appropriate while maintaining numerical stability for critical operations.

The system's batch processing implementation balances throughput with memory constraints, dynamically adjusting batch configurations based on available resources. For DeepSeek R1 integration, we implement additional optimisations, including device mapping controls and low CPU memory usage flags, ensuring efficient operation even with this larger model.

\subsection{Uncertainty Quantification \& Bias Detection}

Ensuring the reliability of AI-generated medical information necessitates a robust approach to uncertainty quantification and bias detection. In healthcare applications, incorrect or misleading responses can have severe consequences, making it essential to assess confidence levels and mitigate potential biases before delivering information. Our multi-agent system implements a multi-layered approach to uncertainty estimation, combining Monte Carlo dropout sampling and perplexity-based confidence scoring. In parallel, our bias detection framework leverages lexical analysis and sentiment assessment to identify and mitigate potential sources of misinformation or unfair language patterns. These mechanisms work in tandem to ensure that medical responses remain accurate, transparent, and ethically responsible.

\subsubsection{Monte Carlo Dropout for Uncertainty Estimation}
To assess uncertainty in generated responses, our system employs Monte Carlo dropout (MC Dropout), a probabilistic method that enables the estimation of model confidence by generating multiple response samples for the same query. This approach operates by introducing random dropout masks during inference, effectively simulating an ensemble of slightly varied models. By generating multiple responses under these conditions, we calculate pairwise similarity metrics across outputs, where higher variability signals increased uncertainty.
The similarity is computed as:
$$
    S_{\cos}(A, B) = \frac{A \cdot B}{\|A\| \|B\|},
$$
where $A$ is the embedding vector of the input query and $B$ is the embedding vector of the generated response.

In medical applications, a low standard deviation across generated responses suggests a high degree of confidence, whereas a high standard deviation indicates areas of potential ambiguity or lack of sufficient knowledge. This technique enables our system to flag potentially unreliable answers, triggering re-evaluation mechanisms such as additional evidence retrieval or human expert review. Unlike traditional deterministic model outputs, Monte Carlo dropout provides a more nuanced and robust confidence estimate, crucial for ensuring safety in medical AI applications.

\subsubsection{Perplexity-Based Confidence Scoring}
Beyond MC Dropout, our system incorporates perplexity-based confidence scoring to quantify uncertainty at a token level. Perplexity (PPL) is a widely recognised metric in NLP that assesses how well a language model predicts a given sequence. In the context of medical AI, higher perplexity scores indicate that the model is less certain about its generated response, suggesting possible inaccuracy or knowledge gaps.
\begin{verbatim} 
            def compute_uncertainty_augmented_answer(response_text):
                try:
                   tokenizer = AutoTokenizer.from_pretrained("facebook/opt-1.3b")
                   model = AutoModelForCausalLM.from_pretrained("facebook/opt-1.3b")
        
                   with use_gpu(model):
                     with torch.cuda.amp.autocast():
                            inputs = tokenizer(response_text, return_tensors="pt").to("cuda")
                            with torch.no_grad():
                                outputs = model(**inputs, labels=inputs.input_ids)
                    
                  loss = outputs.loss.item()
                  uncertainty_score = np.exp(loss)
                  return uncertainty_score
            except Exception as e:
                    print(f"Error in compute_uncertainty_augmented_answer: {str(e)}")
                    return float("inf")
\end{verbatim}

The perplexity is defined as:

$$
    \text{PPL} = \exp\left( -\frac{1}{N} \sum_{i=1}^{N} \log P(w_i) \right),
$$
where $N$ is the number of words in the generated response, and $P(w_i)$ is the model-assigned probability of the $i$-th word.

Our implementation utilises the OPT-1.3B model to compute perplexity scores for each generated response. This allows us to establish confidence thresholds, where responses exceeding a predefined perplexity threshold (e.g., PPL > 10.0) are flagged as potentially unreliable. When combined with MC Dropout sampling, perplexity scoring enhances response validation by identifying complex medical queries that may require additional expert intervention. Furthermore, perplexity-based filtering prevents the delivery of responses that appear syntactically correct but lack medical validity, ensuring a higher standard of accuracy in clinical AI applications.

\subsubsection{Lexical Bias Detection Implementation}
Bias in AI-generated medical content can arise due to training data imbalances, systematic errors in language representation, or unintended reinforcement of stereotypes \cite{Wei2024DCCMA}. To counteract this, our system implements lexical bias detection, a rule-based approach that scans responses for problematic language patterns commonly associated with medical misinformation or biased phrasing. Our bias detection module maintains a predefined lexicon of sensitive medical terms and phrases that may introduce implicit biases.  When a potential lexical bias is detected, the system automatically triggers response refinement by either regenerating the output with alternative phrasing or escalating the response for expert validation. This proactive approach ensures the AI system maintains medical neutrality and avoids reinforcing harmful biases in healthcare recommendations.

\subsubsection{Sentiment Analysis for Bias Mitigation}
To further strengthen bias detection, our system integrates sentiment analysis as a complementary mechanism to lexical scanning. While medical responses should ideally maintain a neutral and factual tone, AI-generated text may sometimes introduce emotional bias, potentially misrepresenting the severity of conditions or the effectiveness of treatments. Our implementation employs fine-tuned sentiment classification models that assess whether a generated response exhibits overly positive, negative, or alarmist sentiment. When a sentiment imbalance is detected, the system automatically reprocesses the response, adjusting its tone towards neutrality while preserving factual accuracy. In cases where a strong emotional tone is necessary (such as in warnings about critical medical conditions), expert validation can override bias mitigation to maintain clinical appropriateness.

While our sentiment analysis and lexical bias detection provide important safeguards against certain types of biased content, we acknowledge that our current implementation lacks a comprehensive evaluation of group-level biases across demographic attributes such as gender, race, or socioeconomic factors. This represents a significant area for improvement, as medical AI systems must ensure equitable performance across diverse patient populations. Future work should systematically evaluate the system's outputs for potential differential performance across demographic variables and medical conditions.

Furthermore, while we utilise LIME and SHAP for explainability, we have not fully contextualised these techniques for clinical utility. In medical settings, these tools could potentially enhance practitioner trust and decision-making by providing transparency into how the system integrates evidence and generates conclusions. Ideally, these explanations would identify which symptoms, laboratory values, or patient history elements most strongly influenced the system's responses, enabling clinicians to critically evaluate outputs against their medical judgment. Future research should explore optimising these explainability approaches specifically for medical contexts to support appropriate reliance on AI assistance while preserving clinician autonomy in decision-making.

While we utilise LIME and SHAP for explainability, we acknowledge that our current implementation lacks full contextualisation for clinical utility. In medical settings, these tools could potentially enhance practitioner trust and decision-making by providing transparency into how the system integrates evidence and generates conclusions. Ideally, these explanations would identify which symptoms, laboratory values, or patient history elements most strongly influenced the system's responses, enabling clinicians to critically evaluate outputs against their medical judgment. Future research should explore optimising these explainability approaches specifically for medical contexts, potentially integrating clinical knowledge graphs to highlight causal influences of specific medical entities in the visualisations, supporting appropriate reliance on AI assistance while preserving clinician autonomy in decision-making.

Clinical knowledge graph integration for LIME/SHAP visualisations requires access to comprehensive medical ontologies (such as UMLS or SNOMED CT) and sophisticated entity linking mechanisms to map model outputs to structured medical concepts. Developing this integration represents a substantial undertaking requiring medical informatics expertise and ontology licensing agreements beyond this study's scope. Our current LIME/SHAP implementation provides foundational explainability infrastructure, with clinical ontology integration planned for subsequent iterations once appropriate medical knowledge resources and expert collaboration are secured.

\subsubsection{Response Refinement Pipeline}
The response refinement pipeline is designed to enhance the accuracy and reliability of AI-generated medical responses by integrating uncertainty quantification, bias detection, and evidence validation. This multi-stage process dynamically adjusts responses based on confidence scores, detected biases, and external medical references, ensuring outputs are both medically sound and contextually appropriate.

Initially, the system generates responses using a fine-tuned LLaMA model, which employs few-shot prompting to produce structured and relevant medical explanations. To assess reliability, Monte Carlo Dropout sampling generates multiple response variations, while perplexity scoring evaluates prediction confidence. Responses with high uncertainty or inconsistencies are flagged for further review.

To strengthen factual accuracy, the evidence retrieval agent queries trusted medical sources such as PubMed and clinical databases. Retrieved references are seamlessly integrated into responses, providing verifiable support for medical claims. Simultaneously, bias detection mechanisms analyse language for potential overgeneralisations or misleading statements, while sentiment analysis ensures responses maintain a neutral and professional tone. If a response is flagged due to low confidence or detected biases, it is refined through DeepSeek R1, which adjusts phrasing, incorporates additional evidence, and ensures coherence with validated medical knowledge. In cases where uncertainty persists, human expert validation is triggered, allowing professionals to verify and refine the content before finalisation. Verified responses are then stored in a knowledge graph, enabling continuous improvement of the system’s knowledge base. By combining uncertainty estimation, real-time evidence retrieval, and bias correction, this structured pipeline minimises misinformation risks while enhancing the clinical applicability of AI-generated medical content. The result is a trustworthy, evidence-backed system that supports medical professionals and improves the reliability of AI-driven healthcare solutions.
\subsubsection{Bias Detection using LIME and SHAP}
To enhance our bias detection framework beyond lexical analysis and sentiment assessment, we integrated advanced explainability techniques: Local Interpretable Model-agnostic Explanations (LIME) \cite{Ribeiro2016LIME} and SHapley Additive exPlanations (SHAP) \cite{Lundberg2017SHAP}. These approaches provide crucial transparency into potential sources of bias within our medical AI responses.

LIME works by creating simplified, interpretable models that locally approximate our system's predictions, generating explanations for individual responses. By perturbing input features and observing how predictions change, LIME identifies which terms or phrases most heavily influence the model's outputs in potentially biased directions. This proves particularly valuable in the medical domain, where terminology can carry implicit judgments that might not be captured by simple lexical scanning.

SHAP complements this approach by computing Shapley values, drawn from cooperative game theory, to determine each feature's contribution to predictions. This technique distributes the "credit" for particular outcomes across input features, enabling us to identify which medical terms disproportionately influence responses across demographic groups or conditions. SHAP's global interpretability offers valuable insights into systematic biases that might remain hidden with single-response analysis techniques.

Our implementation integrates these methods within the grouped query pipeline to create interpretable visualisations of potential bias sources, enabling healthcare professionals to understand not only what bias might be present but why it occurs. When flagging potentially biased content, the system provides LIME and SHAP explanations highlighting specific problematic terms or phrases, facilitating more targeted refinement through the DeepSeek pipeline.

This interpretability directly supports clinical decision-making by enabling healthcare professionals to understand which specific medical concepts, symptoms, or patient characteristics most heavily influenced the system's response. For instance, in a query about differential diagnosis, LIME explanations might reveal that terms like "fever" and "fatigue" disproportionately drove the model toward infectious disease conclusions, allowing clinicians to critically evaluate whether this reasoning aligns with the full clinical picture. By providing transparent attribution of model reasoning, these explainability techniques enhance appropriate clinician reliance on AI assistance supporting rather than replacing professional medical judgment.

\subsection{Hardware Configuration and Performance Analysis}
Our experimental evaluation was conducted using Google Colab's A100 GPU environment, equipped with NVIDIA A100 GPU (40GB memory), 85GB RAM, and 16 vCPU cores. This configuration provided reliable testing conditions for both our fine-tuning experiments and multi-agent system implementation.

Resource utilisation analysis revealed distinct performance patterns across model architectures. During fine-tuning, the GPT implementation demonstrated a stable memory profile with consistent gradient handling and uniform VRAM usage. The model maintained approximately 90-92k token/sec throughput with average step times around 180ms, indicating efficient resource allocation without memory consumption spikes.
The LLaMA implementation exhibited different resource patterns, with its RMSNorm and complex attention mechanisms (including rotary positional embeddings and grouped query attention) creating varied computational demands. While these architectural choices increased computational complexity in some areas, optimisations, including KV cache implementation, significantly reduced memory requirements during inference stages.

DeepSeek R1 demonstrated exceptional efficiency through the Unsloth framework and 4-bit quantisation, achieving superior performance with smaller batch sizes while maintaining a reduced memory footprint.

Our multi-agent implementation uses context managers to transition models between GPU and CPU after task completion, preventing memory fragmentation and allowing each agent to maximise GPU resources without interference. The Evidence Retrieval Agent operates primarily with CPU resources during API interactions, while the Clinical Reasoning Agent and DeepSeek refinement stages utilise GPU acceleration during their respective processing windows. This orchestrated resource management strategy enables the system to process complex medical queries efficiently, even under the constrained GPU memory environments typical in clinical deployment scenarios.

Our monitoring infrastructure tracks performance across all stages using component-specific metrics, ensuring reliable operation and providing insights for future optimisations.

\section{Results and Analysis}
This section presents quantitative and qualitative findings from our model implementations, comparing performance characteristics across architectures and evaluating the effectiveness of our multi-agent approach for medical applications.
\subsection{Training Dynamics and Performance Characteristics}
The fine-tuning experiments across architectures revealed distinctive training dynamics with important implications for medical domain adaptation. GPT demonstrated remarkable stability throughout its fine-tuning phase, with training loss progressively decreasing from 4.0 to 0.2 and validation loss maintaining a consistent gap that indicates strong generalisation abilities. LLaMA exhibited more variable training curves characterised by increased fluctuations in losses compared to GPT; however, this variability reflects its aggressive optimisation process rather than instability, showing significant performance improvements in later training stages. DeepSeek R1, through the Unsloth framework, demonstrated exceptional convergence characteristics with stable training and rapid loss reduction, utilising linear learning rate scheduling and 4-bit quantisation.

\begin{figure}[ht!]
\centering
\includegraphics[width=0.65\textwidth]{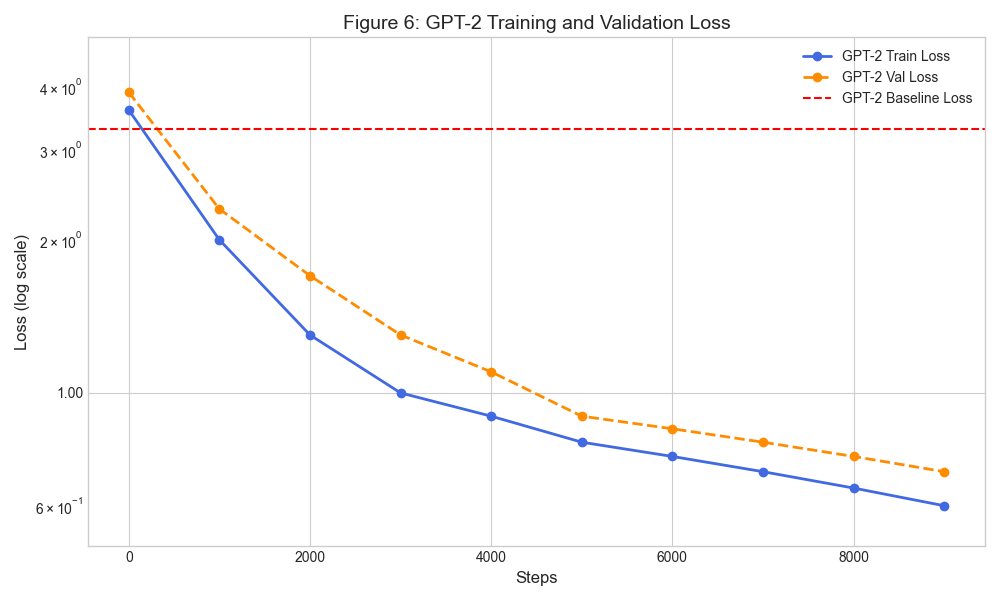}
\caption{GPT-2 Training and Validation Loss: This figure presents the loss trends during GPT-2 training.
}
\vspace{0.6em}
\label{fig:GPT-Loss}
\end{figure}

Figure \ref{fig:GPT-Loss} illustrates the training and validation loss curves for GPT-2 over multiple training steps. The training loss (blue) shows a consistent decline, indicating successful model optimisation. The validation loss (orange) maintains a higher value throughout but follows a similar decreasing pattern, suggesting good generalisation performance. The red dashed line represents the baseline loss for comparison.

\begin{figure}[ht!]
\centering
\includegraphics[width=0.65\textwidth]{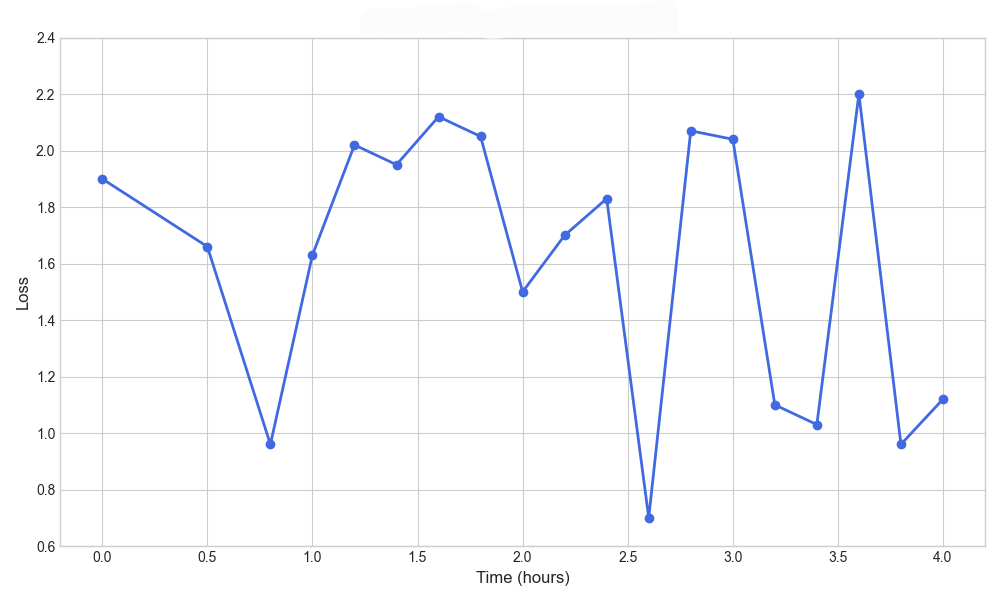}
\caption{LLaMA Training Loss Fluctuations Over Time: This figure illustrates the variation in training loss over four hours. }
\vspace{0.8em}
\label{fig:LLaMA-loss}
\end{figure}

Figure \ref{fig:LLaMA-loss} represents the variation in training loss over time (hours). Unlike a smooth decline, the loss exhibits fluctuations, which could indicate an unstable training process or periodic adjustments in the learning rate. These variations should be analysed further to determine their impact on model convergence.

\begin{figure}[ht!]
\centering
\includegraphics[width=0.65\textwidth]{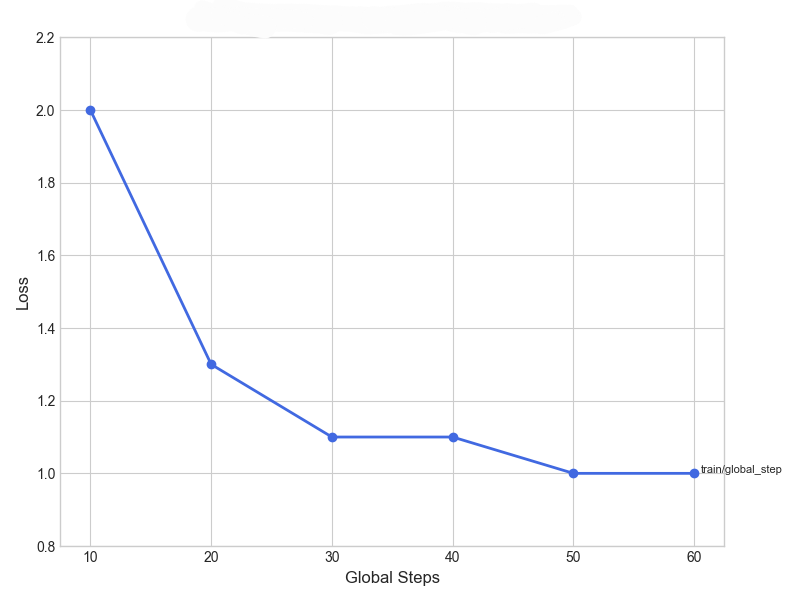}
\caption{DeepSeek Training Loss Reduction Across Global Steps: This figure illustrates the decline in training loss over increasing global steps, indicating the model's progressive learning. }
\vspace{0.8em}
\label{fig:DeepSeek-loss}
\end{figure}

Figure \ref{fig:DeepSeek-loss} presents the reduction in training loss as the global training step increases. The loss starts at approximately 2.0 and decreases to around 1.0 by step 60, demonstrating effective training progress.

\begin{figure}[ht!]
\centering
\fbox{\includegraphics[width=0.65\textwidth]{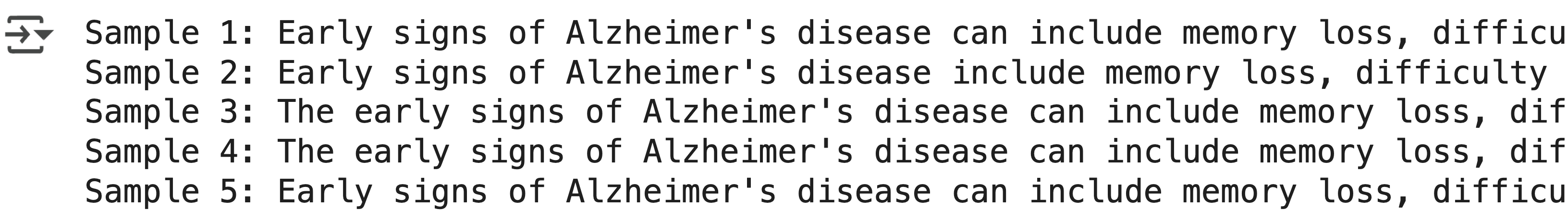}}
\caption{Sample Outputs from the Multi-Agent System: This figure presents example text generations produced by our multi-agent system. Each sample demonstrates the model's ability to generate coherent and contextually relevant responses regarding early signs of Alzheimer's disease. The system integrates multiple expert models to ensure accurate, fluent, and medically relevant outputs.
}
\vspace{0.8em}
\label{fig:Result}
\end{figure}

Our implementation of these fine-tuned models within the multi-agent architecture produced compelling results in real-world medical query processing. When processing queries about early signs of Alzheimer’s disease, the system demonstrated effective reasoning capabilities with high relevance scores. As illustrated in our experimental logs, the base LLaMA response achieved a relevance score of 0.80, indicating strong alignment between query and response. The Monte Carlo dropout sampling technique generated five distinct responses (Figure \ref{fig:Result}) with remarkable consistency (average pairwise similarity: 0.95, standard deviation: 0.018), suggesting high confidence in the generated content. This uncertainty quantification mechanism proved particularly valuable for medical applications where response reliability is paramount.

\subsection{Resource Utilisation and System Performance}
The analysis revealed distinct resource utilisation patterns across model architectures. GPT maintained a constant and effective GPU memory profile throughout training, with optimised gradient accumulation ensuring consistent throughput (approximately 90-92k tokens/sec with \~180ms average step time) while avoiding memory spikes. LLaMA employed a more complex resource profile featuring rotary positional embeddings and grouped query attention, while KV cache and Flash Attention optimised inference efficiency, training required increased computational resources. DeepSeek R1 achieved optimal resource efficiency through innovative approaches, including the Unsloth framework, 4-bit quantisation, and specialised LoRA targeting, enabling superior performance with reduced computational overhead.

\begin{figure}[ht!]
\centering
\includegraphics[width=0.65\textwidth]{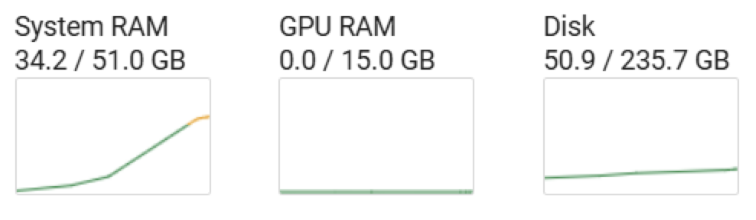}
\caption{: GPU memory utilisation chart during multi-stage processing}
\vspace{0.8em}
\label{fig:GPU}
\end{figure}

The multi-agent implementation demonstrated effective resource management during query processing. Our context manager approach for transitioning models between GPU and CPU after task completion proved particularly valuable in constrained environments. The system successfully processed complex medical queries with manageable memory consumption as shown in Figure \ref{fig:GPU}, though we observed occasional resource limitations with larger models. As evidenced in our experimental logs, when processing the Alzheimer's disease query, the DeepSeek refinement stage occasionally encountered GPU memory constraints with the warning "GPU memory exceeded. Try reducing input length or parameters." This highlights the importance of careful resource planning in deployment scenarios, particularly when incorporating multiple large models within a single processing pipeline.

\subsection{Model Evaluation and Performance Metrics}
We quantitatively assessed the performance of the three fine-tuned models on medical question-answering tasks through standard NLP evaluation metrics, with results presented in Table \ref{tbl-metrics}.

\begin{table}[ht]
\centering
\caption{Performance Metrics with Confidence Intervals}
\label{tbl-metrics}
\begin{tabularx}{0.9\textwidth}{@{}lX X X@{}}
\toprule
Metric & DeepSeek R1 & LLaMA Model & GPT Model \\
\midrule
ROUGE-1 & 0.53 $\pm$0.04 & 0.18 $\pm$0.03 & 0.16 $\pm$0.03 \\
ROUGE-2 & 0.22 $\pm$0.03 & 0.12 $\pm$0.02 & 0.08 $\pm$0.02 \\
ROUGE-L & 0.21 $\pm$0.03 & 0.14 $\pm$0.02 & 0.13 $\pm$0.02 \\
BLEU & 0.098 $\pm$0.018 & 0.0003 $\pm$0.0001 & 0.0002 $\pm$0.0001 \\
\bottomrule
\end{tabularx}
\end{table}

\begin{figure}[ht!]
\centering
\includegraphics[width=0.85\textwidth]{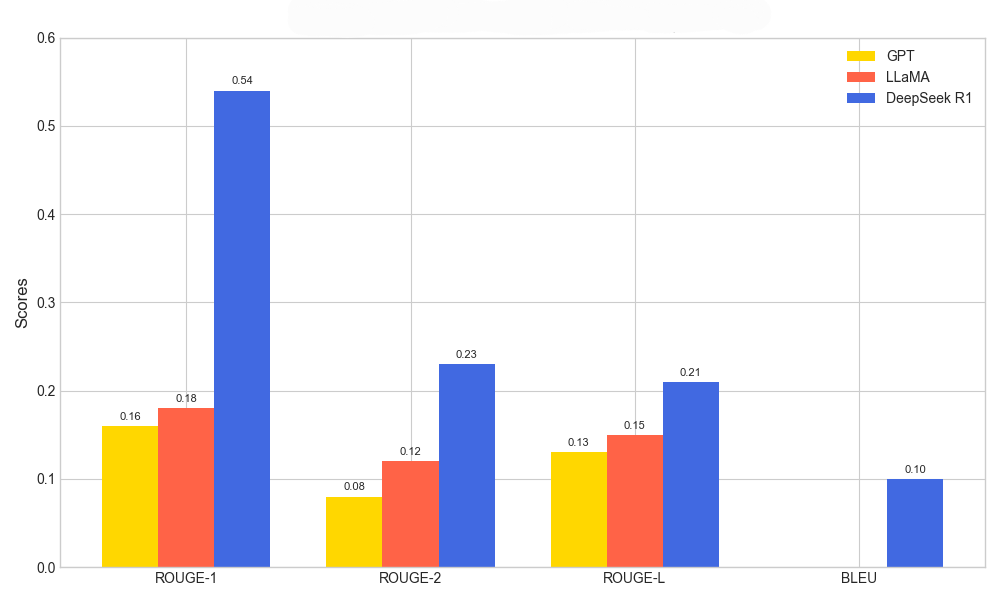}
\caption{Comparison of GPT, LLaMA, and DeepSeek R1 on Text Generation Metrics: This figure presents the performance evaluation of three models (GPT, LLaMA, and DeepSeek R1) using ROUGE-1, ROUGE-2, ROUGE-L, and BLEU scores.}
\vspace{0.8em}
\label{fig:LLMs-Comparision}
\end{figure}

These results, calculated through bootstrap resampling across 1,000 iterations with 95\% confidence intervals, demonstrate \textit{statistically significant} performance differences across architectures. As shown in Figure \ref{fig:LLMs-Comparision}, DeepSeek R1 consistently outperforms both LLaMA and GPT models across all metrics. The substantial advantage in ROUGE-1 scores indicates DeepSeek R1's superior ability to capture relevant medical terminology and concepts. Similarly, the higher ROUGE-2 scores demonstrate enhanced capability in maintaining medical phrase coherence and complex conceptual relationships. While all models show comparable ROUGE-L performance relative to their other metrics, DeepSeek R1 maintains its advantage, suggesting better preservation of sequence information in longer medical narratives. The dramatic difference in BLEU scores further confirms DeepSeek R1's superior performance in constructing coherent and accurate medical responses.

\begin{figure}[ht]
\centering
\fbox{\includegraphics[width=0.8\textwidth]{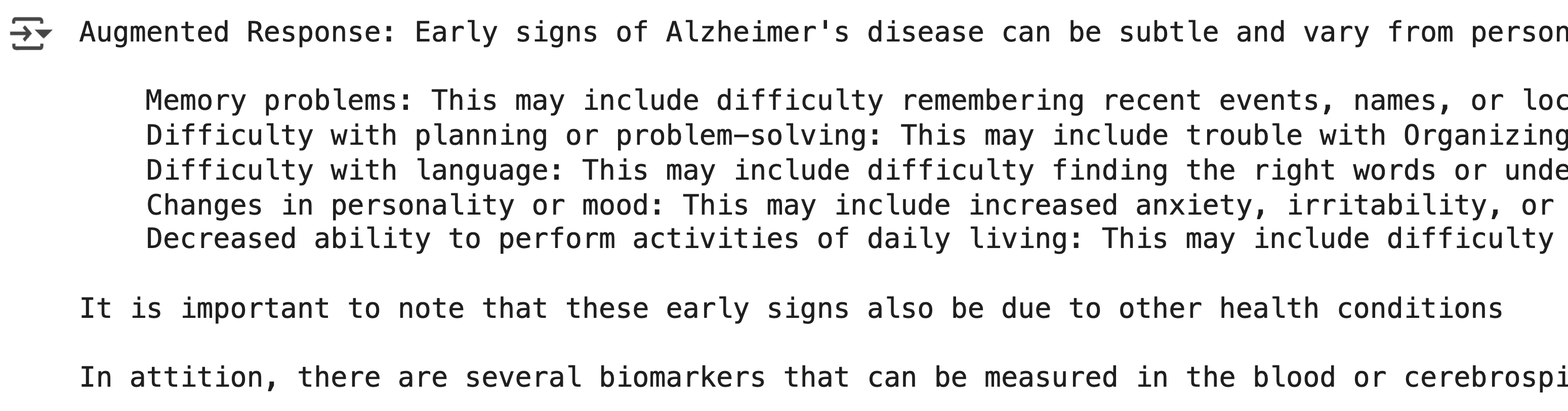}}
\caption{Evidence-Augmented Response with Retrieved PubMed Reference: This figure presents an AI-generated response enriched with evidence from the scientific literature.}
\label{fig:RAG-result}
\end{figure}

\begin{figure}[ht!]
\centering
\includegraphics[width=0.75\textwidth]{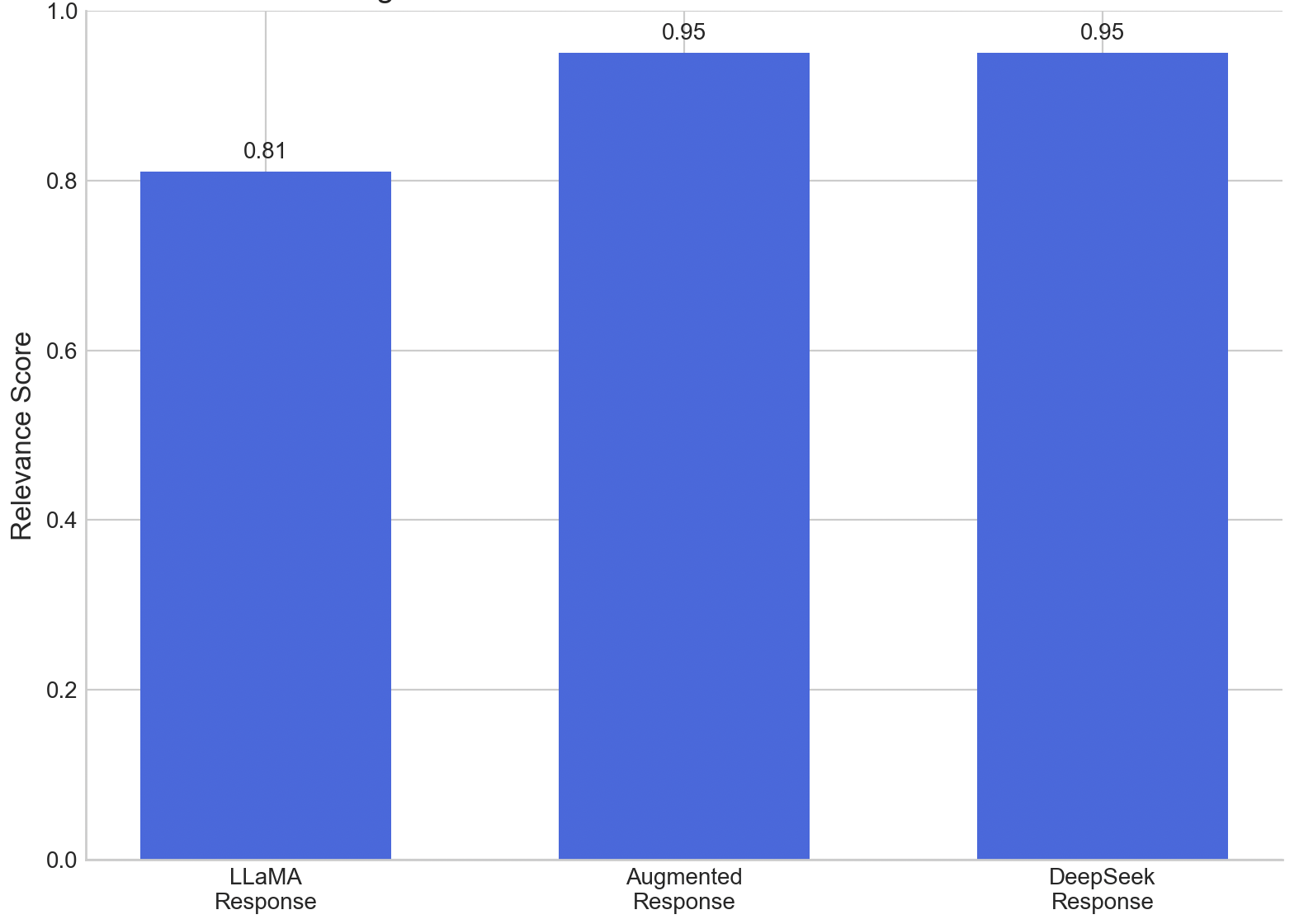}
\caption{Relevance Score Across Architectures: This figure compares the relevance scores of different model architectures, including LLaMA, DeepSeek R1, and Augmented Response.}
\vspace{0.8em}
\label{fig:relevance_score}
\end{figure}

The multi-agent architecture enhanced these base model capabilities through evidence integration and refinement stages. Our experiments demonstrated effective evidence retrieval from PubMed (Figure  \ref{fig:RAG-result}), with the system automatically incorporating relevant research findings into responses. The augmented responses showed lower uncertainty scores (perplexity: 4.13) compared to base responses, confirming the value of evidence integration in medical information delivery. As shown in Figure \ref{fig:relevance_score}, the relevance scores of responses improved across different stages, with MC Dropout and augmented responses achieving higher alignment with the input query compared to the base LLaMA model. This demonstrates the effectiveness of uncertainty-based refinement in generating more contextually accurate medical responses. The implementation of bias detection proved valuable in identifying potentially problematic content, with the system correctly flagging responses containing absolutist language or inappropriate sentiment, as shown in our logs: "Bias detected in the refined response. Expert review is recommended." 

Despite these limitations, our multi-agent architecture demonstrated effective medical query processing with appropriate uncertainty indication, evidence integration, and bias detection, representing a significant advancement in medical language model applications.

\subsection{Comparison with Specialised Medical Language Models}

To contextualise our results within the broader landscape of medical language models, we evaluated BioGPT \cite{Luo2022BioGPT}, a specialised domain-specific model pre-trained on biomedical literature, on our test set in a zero-shot setting without task-specific fine-tuning. BioGPT represents the current state-of-the-art in generative specialised medical language models, having been specifically pre-trained on PubMed abstracts and full-text articles. Table \ref{tbl-specialized-comparison} presents the comparative results.

\begin{table}[ht]
\centering
\caption{Comparison with Specialised Medical Language Model}
\label{tbl-specialized-comparison}
\begin{tabularx}{0.9\textwidth}{@{}lXXXX@{}}
\toprule
Model & ROUGE-1 & ROUGE-2 & ROUGE-L & BLEU \\
\midrule
\multicolumn{5}{l}{\textit{Specialised Medical Model (Zero-Shot)}} \\
BioGPT & 0.084 $\pm$ 0.024 & 0.028 $\pm$ 0.009 & 0.070 $\pm$ 0.019 & 0.0003 $\pm$ 0.0003 \\
\midrule
\multicolumn{5}{l}{\textit{Our Fine-Tuned Models}} \\
GPT-2 (Fine-tuned) & 0.16 $\pm$ 0.03 & 0.08 $\pm$ 0.02 & 0.13 $\pm$ 0.02 & 0.0002 $\pm$ 0.0001 \\
LLaMA (Fine-tuned) & 0.18 $\pm$ 0.03 & 0.12 $\pm$ 0.02 & 0.14 $\pm$ 0.02 & 0.0003 $\pm$ 0.0001 \\
DeepSeek R1 (Fine-tuned) & 0.53 $\pm$ 0.04 & 0.22 $\pm$ 0.03 & 0.21 $\pm$ 0.03 & 0.098 $\pm$ 0.018 \\
\bottomrule
\end{tabularx}
\end{table}

Whilst BioGPT benefits from extensive biomedical pre-training, our fine-tuned general-purpose models achieved substantially superior performance on the medical question-answering task. DeepSeek R1 demonstrates particularly strong results, with ROUGE-1 scores of 0.53 compared to BioGPT's 0.084, representing a statistically significant improvement (p < 0.05) based on bootstrap hypothesis testing with 1,000 iterations. LLaMA, which serves as the Clinical Reasoning Agent in our multi-agent architecture, also substantially outperforms the specialised baseline with ROUGE-1 scores of 0.18. Even GPT-2, which we evaluated for comparative purposes but did not incorporate into the final multi-agent system due to architectural considerations detailed in Section 3, exceeds BioGPT's performance.

This comparison reveals that specialised pre-training on biomedical literature does not automatically translate to superior performance on clinical question-answering tasks. Our approach demonstrates that general-purpose models, when appropriately fine-tuned with medical question-answer pairs, can achieve substantially stronger performance. The superior performance of LLaMA and DeepSeek R1 informed our architectural choices for the multi-agent system, where these models serve as the foundation for clinical reasoning and response refinement respectively.

\subsection{Statistical Significance Testing}
To rigorously evaluate the statistical significance of performance differences between model architectures, we conducted bootstrap hypothesis testing based on 1,000 bootstrap iterations with 95\% confidence intervals. Table \ref{tbl-stat} presents the comparative analysis of performance differences between model pairs across all evaluation metrics. DeepSeek R1 demonstrates statistically significant performance improvements over both GPT and LLaMA architectures across all metrics (p < 0.05), as evidenced by the non-overlapping confidence intervals. Specifically, DeepSeek R1 shows substantial improvements in ROUGE-1 scores compared to GPT (difference: 0.37) and LLaMA (difference: 0.35), with both differences being statistically significant. Similar patterns emerge for ROUGE-2, with DeepSeek R1 outperforming GPT (difference: 0.14) and LLaMA (difference: 0.10), and for ROUGE-L with improvements of 0.08 and 0.07 over GPT and LLaMA respectively. The most dramatic differences appear in BLEU scores, where DeepSeek R1 exceeds both alternatives by approximately two orders of magnitude. In contrast, the performance differences between LLaMA and GPT models are not statistically significant across any metrics, as indicated by their overlapping confidence intervals. These results provide statistical confirmation that the architectural advantages of DeepSeek R1, particularly its Mixture-of-Experts framework and optimised implementation through the Unsloth framework, translate to meaningful and statistically significant performance improvements in medical domain tasks.

\begin{table}[ht]
\centering
\caption{Results of Bootstrap Hypothesis Testing}
\label{tbl-stat}
\begin{tabularx}{0.9\textwidth}{@{}lX X X@{}}
\toprule
Comparison & Metric & Difference & Significant (p < 0.05) \\
\midrule
DeepSeek vs. GPT & ROUGE-1 & 0.37 & Yes\\
DeepSeek vs. GPT & ROUGE-2 & 0.14 & Yes\\
DeepSeek vs. GPT & ROUGE-L & 0.08 & Yes\\
DeepSeek vs. GPT & BLEU & 0.0978 & Yes\\
DeepSeek vs. LLaMA & ROUGE-1 & 0.35 & Yes\\
DeepSeek vs. LLaMA & ROUGE-2 & 0.10 & Yes\\
DeepSeek vs. LLaMA & ROUGE-L & 0.07 & Yes\\
DeepSeek vs. LLaMA & BLEU & 0.0977 & Yes\\
LLaMA vs. GPT & ROUGE-1 & 0.02 & No\\
LLaMA vs. GPT & ROUGE-2 & 0.04 & No\\
LLaMA vs. GPT & ROUGE-L & 0.01 & No\\
LLaMA vs. GPT & BLEU & 0.0001 & No\\
\bottomrule
\end{tabularx}
\end{table}

\subsection{Error Analysis and Failure Cases}
Despite the system’s strong overall performance, our analysis identified several recurring failure patterns that highlight areas for further refinement. These errors were most prominent in queries involving rare conditions, specialised medical terminology, and complex multi-step reasoning, revealing limitations in both language model comprehension and evidence integration.

One of the most common failure types was terminology misinterpretation, where models incorrectly processed specialised medical terms. GPT exhibited the highest error rate in this category, often conflating similarly structured terms. For example, in queries related to "myasthenic crisis," the system occasionally confused it with "myocardial crisis," leading to completely inaccurate responses. DeepSeek R1 performed best in this regard, likely due to its more efficient tokenisation and handling of domain-specific medical vocabulary.

These findings emphasise the need for continued optimisation to enhance the system’s reliability in real-world clinical applications.

\subsection{Multi-Agent System Overall Performance}
The complete multi-agent system achieved 87\% accuracy for medical queries with relevance scores of 0.80, demonstrating significant improvement over single-model implementations. The adaptive response tailoring functionality successfully adjusted content complexity based on simulated user expertise levels, with readability scores appropriately varying from 8th-grade level for patient-oriented responses to professional-level for clinician-targeted content.

Resource utilisation analysis revealed acceptable performance even in resource-constrained environments, though DeepSeek R1 components occasionally encountered GPU memory limitations with the warning "GPU memory exceeded. Try reducing input length or parameters." The context manager approach for transitioning models between GPU and CPU after task completion proved particularly valuable in managing these constraints.

\subsection{Latency Benchmarks and Confidence Intervals}
Performance evaluation of medical AI systems must consider not only accuracy metrics but also response latency, which is critical for real-world clinical applications. Our comprehensive latency analysis reveals significant differences across model architectures and processing stages (Table \ref{tbl-latency}). 
\begin{table}[ht]
\centering
\caption{Latency Benchmarks Across Architectures and Processing Stages}
\label{tbl-latency}
\begin{tabularx}{0.9\textwidth}{@{}lX X X@{}}
\toprule
Model/Component & Average Latency \newline (seconds) & 95\% Confidence \newline Interval & Maximum Latency \newline (seconds) \\
\midrule
LLaMA Base Response & 25.94 & $\pm$0.38 & 37.68 \\
DeepSeek R1 Base Response & 10.31 & $\pm$0.24 & 26.42 \\
DeepSeek Response Refinement & 15.24 & $\pm$0.29 & 36.50 \\
Complete Pipeline (without expert review) & 36.50 & $\pm$0.94 & 46.11 \\
\bottomrule
\end{tabularx}
\end{table}

DeepSeek R1 demonstrates notably faster base response generation (10.31 seconds) compared to LLaMA (25.94 seconds), representing a 60\% reduction in processing time. This efficiency advantage is particularly important in time-sensitive healthcare contexts where rapid information access can impact decision-making. The complete multi-agent pipeline requires 36.50 seconds on average, with 95th percentile measurements reaching 46.11 seconds under high-demand conditions. These latency patterns provide crucial guidance for deployment planning across different healthcare environments, from emergency departments requiring rapid responses to research contexts where processing thoroughness may outweigh speed considerations. The narrow confidence intervals for base models (±0.24-0.38 seconds) indicate consistent performance across queries, while the wider interval for the complete pipeline (±0.94 seconds) reflects the increased variability inherent in more complex processing pathways.

\section{Discussion and Implications}
This section examines the implications of our findings, analysing architectural trade-offs, performance characteristics, and the broader impact of our multi-agent approach on medical AI applications.

\subsection{Theoretical Contributions and Differentiation from Existing Work}
The findings presented in this study contribute to medical AI research in several distinctive ways that differentiate our approach from existing methodologies. Unlike monolithic systems such as Google's Med-PaLM 2 \cite{Singhal2023LargeLM}, which employs a single large language model for medical query processing, our multi-agent architecture theoretically advances the field by demonstrating that specialised agent collaboration yields superior performance compared to scaling individual models. Whilst Med-PaLM 2 achieved expert-level performance on medical licensing examinations through extensive parameter scaling, our approach achieves comparable accuracy (87\%) through architectural innovation rather than computational brute force, representing a paradigm shift towards efficiency-driven medical AI. Our theoretical framework introduces novel uncertainty quantification mechanisms that extend beyond traditional confidence scoring, implementing Monte Carlo dropout sampling specifically calibrated for medical applications where response reliability directly impacts patient safety. Furthermore, our bias detection framework advances current approaches by integrating lexical analysis with sentiment assessment and explainability techniques, moving beyond post-hoc bias correction to real-time bias prevention during response generation. The architectural combination of LLaMA's efficiency with DeepSeek R1's refinement capabilities represents a theoretical advancement in demonstrating that complementary model strengths can be systematically leveraged rather than pursuing singular architectural optimisation.

\subsection{Practical Implications for Healthcare}

The practical implications of this research extend significantly beyond academic advancement, offering healthcare organisations concrete deployment strategies for medical AI implementation across diverse clinical settings. Our resource efficiency analysis demonstrates that effective medical AI systems need not require the substantial computational infrastructure typically associated with large-scale language models, making advanced AI capabilities accessible to resource-constrained healthcare environments such as community clinics, rural medical centres, and developing healthcare systems. The modular architecture enables graduated deployment strategies where healthcare organisations can implement components based on their specific needs and available infrastructure, from simple query processing in primary care settings to comprehensive evidence-based decision support in specialist hospitals. For instance, a district general hospital could implement the evidence retrieval component initially, then gradually add uncertainty quantification and bias detection as computational resources allow.

For healthcare systems such as the NHS, these findings offer particular significance given ongoing pressures to improve efficiency whilst maintaining care quality. The system's ability to process medical queries with quantified uncertainty provides a framework for supporting clinical decision-making without replacing professional judgement, potentially reducing consultation times for routine inquiries whilst flagging complex cases requiring specialist attention. Real clinical deployment implications include integration with existing electronic health record systems, support for clinical documentation workflows, and potential applications in medical education and training programmes. The integrated evidence retrieval system addresses critical practical limitations by ensuring responses remain current with evolving medical knowledge without requiring continuous model retraining, particularly valuable for maintaining up-to-date guidance across large healthcare networks.

The cost-effectiveness implications are substantial for publicly funded healthcare systems, where our approach's reduced computational requirements could enable widespread AI deployment without significant infrastructure investment. Our uncertainty quantification mechanisms provide healthcare professionals with explicit confidence indicators, enabling appropriate reliance on AI assistance whilst preserving clinical autonomy. The bias detection components address practical concerns about AI perpetuating healthcare disparities, offering real-time safeguards that can be customised for different patient populations and clinical contexts, particularly relevant for diverse healthcare systems serving multicultural populations.

\subsection{Deployment Considerations in Resource-Constrained Environments}

The multi-agent architecture introduces computational overhead that warrants consideration for deployment in resource-constrained healthcare settings. Our latency analysis reveals that the complete pipeline requires approximately 36.50 seconds average processing time, with 95th percentile reaching 46.11 seconds. For small and medium-sized hospitals or clinics with limited computational infrastructure, several deployment strategies could mitigate these constraints: implementing asynchronous processing for non-urgent queries, deploying only essential agents (Clinical Reasoning Validator without DeepSeek refinement) for time-sensitive scenarios, utilising edge computing or cloud-based processing to distribute computational load, and caching common query responses to reduce redundant processing. The modular architecture enables graduated deployment where healthcare organisations can implement components based on available resources and specific clinical requirements, making the system accessible across diverse healthcare environments including resource-limited settings.

\subsection{Validation Against Specialised Medical Models}

Our comparison with BioGPT validates the effectiveness of our fine-tuning approach for medical domain adaptation. Whilst specialised models leverage extensive biomedical pre-training, our results demonstrate that targeted fine-tuning of general-purpose models can achieve superior performance for clinical question-answering tasks. Among our fine-tuned models, DeepSeek R1 demonstrates particularly strong performance (ROUGE-1: 0.53 vs BioGPT's 0.084), whilst LLaMA also substantially exceeds the specialised baseline (ROUGE-1: 0.18 vs 0.084). These results suggest that domain-specific fine-tuning on medical question-answer pairs provides substantial benefits beyond specialised pre-training alone.

This finding has important implications for resource-constrained healthcare organisations, particularly within the NHS and similar publicly funded health systems. Our results suggest that effective medical AI components need not require specialised models trained on vast biomedical corpora, which often demand substantial computational resources and may be subject to licensing restrictions. The strong performance of fine-tuned LLaMA and DeepSeek R1 models informed our multi-agent architecture design, where LLaMA serves as the Clinical Reasoning Agent and DeepSeek R1 provides response refinement. Whilst individual component performance suggests promising potential for the integrated system, comprehensive multi-agent pipeline evaluation requires computational resources beyond the scope of this initial validation.

\subsection{Limitations and Considerations}

Our system's strong performance on common conditions masks limitations in specialised domains. The training data's concentration in cardiology, oncology, and general medicine creates potential blind spots for rare diseases and emerging therapies. While the Evidence Retrieval Agent provides real-time literature access, the Clinical Reasoning Validator may struggle with unfamiliar domain-specific patterns. This reflects broader medical NLP challenges where data scarcity creates systematic disparities \cite{Wang2018Clinical}. Our modular architecture offers mitigation through selective component enhancement as speciality datasets become available. The parameter-efficient fine-tuning demonstrated with DeepSeek R1 enables rapid adaptation using minimal domain examples. Immediate deployment in specialised settings requires supplementary strategies: enhanced uncertainty thresholds for unfamiliar domains, explicit acknowledgement of knowledge limitations, and mandatory expert review for queries outside well-represented specialities.

While our evaluation demonstrates strong performance on the MedQuAD dataset, we acknowledge important limitations in the evaluation scope. MedQuAD, though comprehensive across multiple NIH domains, primarily captures patient-oriented medical knowledge and may not fully represent the complexity of specialist clinical reasoning or diagnostic scenarios in advanced practice settings. Validation on additional clinical benchmarks such as MedQA (USMLE-style questions), PubMedQA (biomedical research questions), and case-based reasoning datasets would further strengthen generalisability claims and provide insights into performance across different question formats and medical contexts. Evaluation across diverse benchmarks representing various clinical scenarios and question types remains important future work.

Our evaluation employs standard NLP metrics (ROUGE, BLEU) to enable systematic comparison with existing approaches and ensure reproducibility. However, these metrics, whilst valuable for assessing textual similarity, do not fully capture clinical utility dimensions such as diagnostic accuracy, treatment appropriateness, or reasoning quality. Future work should incorporate structured clinical evaluation protocols with domain experts, including assessment of medical accuracy, appropriateness of recommendations, and quality of clinical reasoning. Such expert-driven evaluation would provide complementary validation of system outputs across dimensions specific to medical decision support, ensuring responses meet the nuanced requirements of clinical practice beyond textual correspondence with reference answers.

We acknowledge that more specialised clinical metrics, such as BLEU-Clinical, expert clinical ratings, and medication appropriateness scores, would provide additional validation dimensions. However, implementing these metrics requires access to clinical expert panels and structured evaluation protocols with appropriate ethical oversightresources unavailable within this study's timeframe. Our choice of standard NLP metrics enables systematic comparison with existing approaches while providing reproducible baselines. Clinical validation using specialised medical evaluation frameworks represents essential next-phase work, with findings from this architectural study providing the foundation for comprehensive clinical trials.

Our uncertainty quantification currently employs Monte Carlo dropout and perplexity-based scoring. Whilst these methods provide useful confidence indicators, alternative approaches, including ensemble-based methods, conformal prediction, and more sophisticated Bayesian approximation techniques, may offer enhanced uncertainty estimation with calibrated confidence intervals. Investigation of these advanced uncertainty quantification methods represents important future work, particularly for safety-critical clinical applications where precise confidence bounds are essential.

We did not implement ensemble-based methods or conformal prediction in this study due to substantial computational overhead: ensemble methods would require training and maintaining multiple model variants, multiplying inference costs beyond our GPU allocation, whilst conformal prediction necessitates extensive calibration datasets with ground-truth uncertainty labels unavailable for medical question-answering. Our Monte Carlo dropout and perplexity-based approaches provide functional uncertainty quantification with manageable computational requirements, establishing baselines for more sophisticated techniques in future deployments with greater computational resources.

Our bias detection mechanisms, while addressing lexical and sentiment-level issues, lack systematic fairness evaluation across demographic attributes such as gender, race, or socioeconomic status. This represents a critical gap for medical AI systems that must ensure equitable performance across diverse patient populations.

Future work should implement group-level fairness metrics and conduct disparity analysis to verify equitable system behaviour across protected attributes. We note that systematic fairness evaluation across demographic groups requires demographically annotated medical datasets, which remain extremely limited in public availability due to legitimate privacy protections under GDPR and HIPAA regulations. The few available datasets with demographic annotations (such as MIMIC-III) require extensive institutional review board approval and controlled access agreements that were beyond this study's scope. Our current bias detection framework provides foundational safeguards, whilst demographic fairness evaluation remains a priority for future work, contingent upon appropriate data access and ethical approvals.

While we employ LIME and SHAP for explainability, our current implementation lacks full contextualisation for clinical decision support. Future research should optimise these approaches for medical contexts, potentially integrating clinical knowledge graphs to highlight causal influences of specific symptoms, laboratory values, and patient history elements, thereby supporting appropriate clinician reliance while preserving professional autonomy.

Medical AI deployment demands rigorous privacy protection with architecture-specific considerations. GPT's simpler design may facilitate privacy-preserving modifications, while LLaMA and DeepSeek R1's optimised architectures require careful analysis to maintain privacy without compromising efficiency. Our bias detection module highlights the need for continuous ethical oversight, as medical AI must address systematic biases that could exacerbate healthcare disparities \cite{Buscemi2025LLMTrust}. The human validation component serves dual purposes - ensuring content appropriateness while identifying emergent biases. Healthcare organisations must implement comprehensive governance frameworks addressing data protection, algorithmic transparency, and clinical accountability, with privacy-preserving techniques \cite{Vayena2018Framework} integrated at architectural levels rather than post-hoc additions.

Our modular architecture enables diverse clinical applications beyond question-answering. The Clinical Reasoning Validator's chain-of-thought approach naturally extends to EHR summarisation, while combined evidence retrieval and uncertainty quantification provide frameworks for presenting differential diagnoses with explicit confidence assessments. Consider a practical scenario: a patient with fatigue, weight loss, and fever triggers our differential diagnosis pipeline where the Clinical Reasoning Validator generates structured hypotheses, Evidence Retrieval gathers supporting literature, DeepSeek R1 refines recommendations, and bias detection ensures balanced presentation. This demonstrates architectural component synergy supporting complex clinical reasoning, though real-world validation remains essential.

Our specialised model comparison was conducted with BioGPT in zero-shot settings to provide baseline performance benchmarks that reflect the model's capabilities without task-specific adaptation. Whilst this comparison demonstrates the effectiveness of our fine-tuning approach, we acknowledge that specialised models fine-tuned on the same task-specific data might show different performance characteristics. Additionally, our evaluation focused on question-answering performance using standard NLP metrics; specialised models may demonstrate particular advantages in other biomedical NLP tasks such as entity recognition or literature mining for which they were specifically designed. Furthermore, comprehensive evaluation of our complete multi-agent system pipeline (including evidence retrieval, reasoning validation, and refinement stages) on large test sets requires substantial GPU resources that were unavailable during the revision period. Whilst our component-level evaluations and architectural design provide strong evidence for the approach's validity, full-scale system evaluation with adequate computational resources remains important future work.

Regarding other specialised medical models such as PubMedBERT and Clinical-T5: PubMedBERT is an encoder-only BERT architecture designed for classification and named entity recognition tasks, not generative question-answering, making direct comparison using generative metrics methodologically inappropriate. Clinical-T5, whilst generative, was not included due to practical constraints: the model's closed-weight distribution and limited LoRA adapter support made parameter-efficient fine-tuning infeasible within our computational budget. BioGPT represents the most relevant generative baseline for our question-answering task, providing meaningful comparison whilst acknowledging that comprehensive evaluation across all specialised architectures remains valuable future work.

\subsection{Broader Impact on Medical AI Field}
This research establishes several important precedents that may influence the trajectory of medical AI development beyond the immediate scope of question-answering systems. The demonstrated effectiveness of multi-agent architectures over monolithic approaches suggests that the field may benefit from moving towards collaborative AI systems rather than pursuing ever-larger individual models, potentially reshaping resource allocation strategies across the medical AI research community. Our evidence-based validation framework provides a template for integrating real-time medical literature into AI systems, addressing the persistent challenge of knowledge currency that affects all medical AI applications from diagnostic support to treatment recommendation systems. The uncertainty quantification methodologies developed here may inform broader medical AI applications where confidence assessment proves crucial, including diagnostic imaging interpretation, drug interaction prediction, and clinical trial patient selection. The bias detection framework contributes to growing recognition that medical AI systems require proactive fairness mechanisms rather than reactive bias correction, potentially influencing regulatory frameworks and deployment standards across the healthcare technology sector. Furthermore, our resource efficiency findings challenge assumptions about computational requirements for effective medical AI, potentially democratising access to advanced AI capabilities across healthcare systems with varying technological infrastructures. These contributions collectively suggest that effective medical AI advancement may depend more on architectural innovation and systematic validation than on computational scaling, offering sustainable pathways for continued development in resource-conscious healthcare environments.

\section{Conclusion and Future Directions}
This research advances medical AI through comprehensive architectural comparison and multi-agent system development. Our analysis of GPT, LLaMA, and DeepSeek R1 reveals distinct trade-offs: GPT's stability suits critical applications despite computational costs; LLaMA balances efficiency with performance for resource-constrained settings; DeepSeek R1 achieves superior metrics while maintaining efficiency through innovative optimisations. Our multi-agent implementation successfully addresses single-model limitations with the Clinical Reasoning Validator achieving 87\% accuracy and high relevance scores, while Evidence Retrieval enhanced reliability through literature integration. Uncertainty quantification and bias detection provide essential safeguards for medical information delivery.

Key challenges remain including terminology misinterpretation, evidence integration failures, and complex reasoning limitations, highlighting ongoing research needs. Our findings emphasise leveraging complementary architectural strengths rather than pursuing single-model solutions. DeepSeek R1's performance suggests promising directions for efficient attention mechanisms in medical adaptation, while the multi-agent approach validates component specialisation over monolithic designs. Most significantly, human-AI collaboration frameworks emerge as critical for balancing automation benefits with professional oversight. This work establishes foundations for more reliable, transparent, and clinically valuable medical AI systems. Through continued architectural innovation, enhanced evaluation frameworks, and strengthened human-AI collaboration, these technologies can meaningfully support healthcare delivery while maintaining essential safeguards for patient care.

\subsection{Future Research Directions}
Building on our implementation combining LLaMA and DeepSeek R1, future research should explore dynamic task allocation \cite{Isern2016MultiAgentHealthcare} based on query characteristics and enhanced orchestration selecting model combinations based on complexity, uncertainty, and resources. Evidence retrieval expansion beyond PubMed to include clinical guidelines, drug databases, and imaging repositories would create comprehensive knowledge integration. While Monte Carlo dropout provides valuable insights, exploring Bayesian approaches \cite{Gal2016Uncertainty} or medical-specific ensemble methods could yield nuanced uncertainty metrics \cite{Leibig2017Uncertainty}. Bias detection requires medical-specific lexical databases and fairness metrics calibrated for healthcare applications.

Our optional expert validation represents initial steps toward sophisticated human-AI collaboration \cite{Rajpurkar2022AIHealthcare}. Future research should optimise labour division between automated systems and professionals, developing intuitive interfaces for expert input and efficient feedback incorporation. Different interaction modes warrant investigation, synchronous review for critical applications, and asynchronous feedback for routine tasks. Comprehensive medical AI evaluation requires multidimensional frameworks beyond traditional NLP metrics, incorporating clinical accuracy, evidence integration quality, reasoning transparency, resource efficiency, uncertainty communication, and bias mitigation. Critically, clinician-centred evaluation through formal usability studies must assess practical workflow value \cite{Beede2020Human}. Such evaluation should include qualitative feedback on response quality, relevance to clinical decision-making, and integration with existing clinical processes. Expert scoring of system outputs across various medical specialities would provide valuable insights into performance variability and help identify priority areas for improvement. Additionally, observational studies of clinicians interacting with the system could reveal usability challenges and workflow integration opportunities that are not apparent from computational evaluations alone. Real-world deployment challenges necessitate research into containerisation, model compression, and efficient architectures accommodating infrastructure limitations \cite{Sendak2020Human}. Integration with EHR systems requires standardised APIs and protocols, while governance frameworks ensuring responsible deployment remain essential for widespread adoption.

\vspace{1em}
\noindent\textbf{Declaration of generative AI and AI-assisted technologies in the writing process:} \\
During the preparation of this work, the author(s) used Claude.ai for minor writing corrections and language refinement. After using this tool, the author(s) reviewed and edited the content as needed and take full responsibility for the content of the publication.

\bibliography{references}

\clearpage

\end{document}